\documentclass[journal]{IEEEtran}
\usepackage{graphicx}
\usepackage{bm}
\usepackage{amsmath}
\usepackage[numbers,sort&compress]{natbib}
\usepackage{algorithm}
\usepackage{color}
\usepackage{amsfonts}
\usepackage{amsmath}
\usepackage{multirow}
\usepackage{threeparttable} 
\usepackage{chngpage}
\usepackage{epstopdf}
\usepackage{epsfig}
\usepackage{Extpfeil}
\usepackage[table]{xcolor}
\usepackage{fixltx2e}
\usepackage{bbm}
\usepackage{amsmath}
\usepackage{amsfonts}
\usepackage{algorithm}
\usepackage{caption}
\usepackage{bm}
\usepackage[table]{xcolor}
\usepackage{color}
\usepackage{algorithmic}
\usepackage{epstopdf} 
\usepackage{epsfig}
\newcommand{\argmin}{\operatornamewithlimits{argmin}}
\usepackage{graphicx}

\begin{document}
\title{Clustering Ensemble  Meets Low-rank Tensor Approximation}

\author{Yuheng~Jia, Hui~Liu, Junhui~Hou,
	and Qingfu~Zhang
	\thanks{This work was supported by the Hong Kong Research Grants Council under Grants 21211518 and 11219019.}
	\thanks{Y. Jia is with the School of Computer Science and Engineering, Southeast
		University, Nanjing, 211189, China. Part of this work was finished when he was with  the Department
		of Computer Science, City University of Hong Kong, Kowloon, Hong Kong, SAR.  (e-mail: yhjia@seu.edu.cn). }
	\thanks{
		H. Liu, J. Hou ans Q. Zhang are with the Department
		of Computer Science, City University of Hong Kong, Kowloon, Hong Kong, SAR. 
		(e-mail: hliu99-c@my.cityu.edu.hk; jh.hou@cityu.edu.hk; qingfu.zhang@cityu.edu.hk ).} 
}
\maketitle
\IEEEpeerreviewmaketitle

\begin{abstract}
	This paper explores the problem of clustering ensemble, which aims to combine multiple base clusterings to produce better performance than that of the individual one. The existing clustering ensemble methods generally construct a co-association matrix, which indicates the pairwise similarity  between samples, as the weighted linear combination of the connective matrices from different base clusterings, and the resulting co-association matrix is then adopted as the input of an off-the-shelf clustering algorithm, e.g., spectral clustering. However, the co-association matrix may be dominated by poor base clusterings, resulting in inferior performance. In this paper, we propose a novel low-rank tensor approximation-based method to solve the problem from a global perspective. Specifically, by inspecting whether two samples are clustered to an identical cluster under different base clusterings, we derive a coherent-link matrix, which contains limited but highly reliable relationships between samples. We then stack the coherent-link matrix and the co-association matrix to form a three-dimensional tensor, the low-rankness property of which is further explored to propagate the information of the coherent-link matrix to the co-association matrix, producing a refined co-association matrix. We formulate the proposed method as a convex constrained  optimization problem and solve it efficiently. Experimental results over 7 benchmark data sets show that the proposed model achieves a breakthrough in clustering performance, compared with 12 state-of-the-art methods.  To the best of our knowledge, this is the
	first work to explore the potential of low-rank tensor on clustering
	ensemble, which is fundamentally different from previous approaches.  
\end{abstract}

\section{Introduction}
\noindent Clustering is an important but very challenging unsupervised task, the goal of which   is to partition a set of samples into homogeneous groups \cite{8486482}. Numerous applications can be formulated as a clustering problem, such as recommender systems \cite{song2014online}, community detection \cite{wu2018nonnegative}, and image segmentation \cite{li2019superpixel}. Over the past decades,  a  large number of clustering techniques were proposed,  e.g., K-means \cite{jain2010data}, spectral clustering \cite{vonLuxburg2007}, matrix factorization \cite{jia2019semi,8361078,9013063}, hierarchical clustering \cite{johnson1967hierarchical}, Gaussian mixture models \cite{moore1999very}, and so on. As each method has its own advantages as well as drawbacks, no method could always outperform others \cite{vega2011survey}. Additionally, a clustering method usually contains  a few hyper-parameters, on which its performance heavily depends \cite{9178787}. 
Moreover, the hyper-parameters are difficult to tune, and some methods are quite sensitive to initialization, like K-means. Those dilemmas increase the difficulty in choosing an appropriate clustering method for a typical clustering task. To this end,  clustering ensemble was introduced, i.e., given a set of base clusterings produced by different methods or the same method with different hyper-parameters/initializations,  it aims to generate a consensus clustering  with better clustering performance than the base clusterings \cite{sagi2018ensemble,boongoen2018cluster}.
Unlike supervised ensemble learning, clustering ensemble is more difficult \cite{tao2017ensemble,tao2016robust}, as the commonly used strategies in supervised ensemble learning, such as voting, cannot be directly applied to clustering ensemble, when labels of 
samples are unavailable. 

To realize clustering ensemble, the existing methods generally first  learn a pairwise relationship matrix from the base clusterings, and then apply off-the-shelf clustering methods like spectral clustering to the resulting matrix to produce the final clustering result \cite{tao2017simultaneous}. Based on how to generate the pairwise relationship matrix, we roughly divide the existing methods into two categories. 
($1$) The first kind of methods treats the base clusterings as new feature representations (as shown in Fig. 1-A), to learn a pairwise relationship matrix. 
For example, \cite{gao2016robust} formulated clustering ensemble  as a convex low-rank matrix representation problem. \cite{zhou2019ensemble} used a Frobenius norm regularized self-representation model to seek a dense affinity matrix for clustering ensemble. 
($2$) The second kind of methods relies on the co-association matrix (as shown in Fig. 1-C), which summarizes the co-occurrence of samples in the same cluster of the base clusterings. The concept of using co-association matrix was first proposed by \cite{fred2005combining}, and since then it became popular as an important fundamental method in clustering  ensemble. \cite{7811216} theoretically bridged the co-association based method  to weighted K-means clustering, which largely reduces the computational complexity. 
Recently, many advanced co-association matrix construction methods were proposed. For example, \cite{huang2017locally}  considered the uncertainty of each base clustering and proposed a locally weighted co-association matrix. \cite{huang2018enhanced} used the cluster-wise similarities to enhance the traditional co-association matrix. \cite{zhou2020self} proposed a self-paced strategy to learn the co-association matrix.  See the detailed discussion about the related works in the next section. 
We observe that, the constructed co-association matrices of the prior works are  variants of a weighted linear combination of the connective matrices (as shown in Fig. 1-B) from different base clusterings. When the performance of some base clusterings are poor, they will dominate the co-association matrix and degrade the clustering performance severely. 

In this paper, we propose a novel constrained low-rank tensor approximation (LTA) model to refine the co-association matrix from a global perspective. 
Specifically, as shown in Fig. 1-D, we first construct a coherent-link matrix, whose element examines whether two samples are from the same cluster in all the base clusterings or not. 
We then stack the coherent-link matrix and the conventional co-association matrix to form a $3$-dimensional (3-D) tensor shown in Fig. 1-E, which is further low-rank approximated.
By exploring the low-rankness,  the proposed model can propagate the highly reliable information of  the coherent-link matrix to the co-association matrix, producing a refined co-association matrix,  
which is adopted as the input of an  off-the-shelf  clustering method to produce the final clustering result.
Technically, the proposed model is formulated as a convex optimization problem and solved by an alternative iterative method. 
We evaluate the proposed model on $7$ benchmark data sets, and compare it with $12$ state-of-the-art clustering ensemble methods. The experimental comparisons substantiate that  the proposed model significantly outperforms state-of-the-art methods. 
\textit{To the best of our knowledge, this is the first work to explore the potential of low-rank tensor on clustering ensemble. }


\begin{figure}[!t]
	\centering
	\centerline{\epsfig{figure=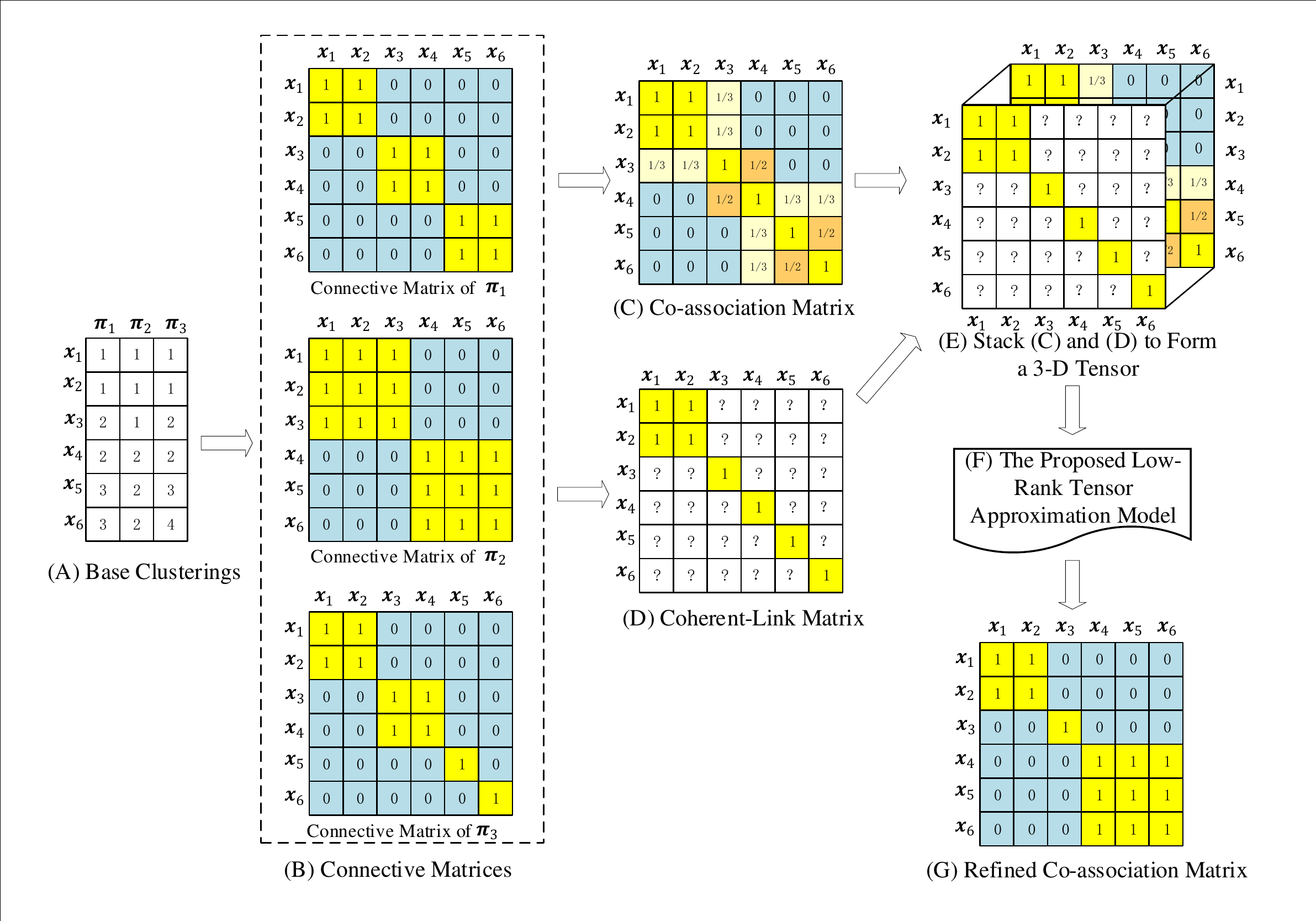,width=8.5cm}}
	\caption{Illustration of the proposed method by taking $3$ base clusterings denoted by $\mathbf{\pi}_1$, $\mathbf{\pi}_2$ and $\mathbf{\pi}_3$,  and $6$ input samples denoted by $\mathbf{x}_1,\cdots,\mathbf{x}_6$ as an example. By exploring the low-rankness of the formed $3$-D tensor, the limited but highly reliable information contained in the coherent-link matrix can be leveraged to enhance the quality of the co-association matrix.}
	\label{fig:framework-s}
\end{figure}

\section{Related Work}
\subsubsection{Notation.}
We denote tensors by boldface swash letters, e.g., $\bm{\mathcal{A}}$, matrices by boldface capital letters, e.g., $\mathbf{A}$, vectors by boldface lowercase letters, e.g., $\mathbf{a}$, and scalars by lowercase letters, e.g., $a$. 
Let $\bm{\mathcal{A}}(i,j,k)$ denote the $(i,j,k)$-th element of $3$-D tensor $\bm{\mathcal{A}}$, $\mathbf{A}(i,j)$ denote the $(i,j)$-th element of matrix $\mathbf{A}$, and $\mathbf{a}(i)$ denote the $i$-th entry of vector $\mathbf{a}$. The $i$-th frontal slice of tensor $\bm{\mathcal{A}}$ is denoted as $\bm{\mathcal{A}}(:,:,i)$.

\subsubsection{Rank of tensors.}
In this paper, we use the tensor nuclear norm induced by tensor singular value decomposition (t-SVD) \cite{kilmer2013third} to measure  the rank of a tensor. Specifically, the t-SVD of a $3$-D tensor $\bm{\mathcal{A}}\in\mathbb{R}^{n_1\times n_2\times n_3}$ can be represented as
\begin{equation}
\bm{\mathcal{A}}=\bm{\mathcal{U}}* \bm{\mathcal{S}} * \bm{\mathcal{V}}^\mathsf{T},
\label{t-SVD-form}
\end{equation}
where $\bm{\mathcal{U}}\in\mathbb{R}^{n_1 \times n_1 \times n_3}$ and $\bm{\mathcal{V}}\in\mathbb{R}^{n_2\times n_2\times n_3}$ are two orthogonal tensors, $\bm{\mathcal{S}}\in\mathbb{R}^{n_1\times n_1\times n_3}$ is an f-diagonal tensor, 
$*$ and $\cdot^\mathsf{T}$ denote tensor product and tensor transpose, respectively. The detailed definitions of the above-mentioned tensor related operators can be find in \cite{zhang2014novel}.  Since the tensor product can be efficiently computed in the Fourier domain \cite{kilmer2013third}, the t-SVD form of a tensor can be obtained with fast Fourier transform (FFT) efficiently  as shown Algorithm 1. 
Given t-SVD, the tensor nuclear norm \cite{zhang2014novel} is defined as the sum of the absolute values of the diagonal entries of $\bm{\mathcal{S}}$, i.e.,  
\begin{equation}
\|\bm{\mathcal{A}}\|_{\star} =\sum_{i=1}^{{\rm min}(n_1,n_2)} \sum_{k=1}^{n_3}|\bm{\mathcal{S}}(i,i,k)|.
\end{equation}

\begin{algorithm}[!t]
	\caption{t-SVD of a $3$-D tensor \cite{zhang2014novel}}
	\begin{algorithmic}[1]
		\renewcommand{\algorithmicrequire}{\textbf{Input:}}
		\renewcommand{\algorithmicensure}{\textbf{Initialize:}}
		\REQUIRE  $3$-D tensor  $\bm{\mathcal{A}}\in\mathbb{R}^{n_1 \times n_2 \times n_3}$. 
		\STATE Perform FFT on $\bm{\mathcal{A}}$, i.e., $\bm{\mathcal{A}}_f={\rm fft}(\bm{\mathcal{A}},[~],3)$;
		\STATE \textbf{for} $k=1:n_3$; \textbf{do}
		\STATE ~~Perform SVD on each frontal slice of $\bm{\mathcal{A}}_f$, i.e., ~~[$\mathbf{U}$,$\mathbf{S}$,$\mathbf{V}$]=SVD($\bm{\mathcal{A}}_f(:,:,k)$) ;
		\STATE ~~~$\bm{\mathcal{U}}_f(:,:,k)=\mathbf{U}$, $\bm{\mathcal{S}}_f(:,:,k)=\mathbf{S}$, $\bm{\mathcal{V}}_f(:,:,k)=\mathbf{V}$;
		
		\STATE \textbf{end}
		\STATE Perform inverse FFT on $\bm{\mathcal{U}}_f$, $\bm{\mathcal{S}}_f$ and $\bm{\mathcal{V}}_f$, i.e.,
		$\bm{\mathcal{U}}=\rm{ifft}(\bm{\mathcal{U}}_f, [ ],3)$, $\bm{\mathcal{S}}=\rm{ifft}(\bm{\mathcal{S}}_f, [ ],3)$ and $\bm{\mathcal{V}}=\rm{ifft}(\bm{\mathcal{V}}_f, [ ],3)$;
	\end{algorithmic}
	\textbf{Output:} $\bm{\mathcal{U}}$, $\bm{\mathcal{S}}$ and $\bm{\mathcal{V}}$.
	\label{t-SVD}
\end{algorithm}

\subsubsection{Formulation of  Clustering Ensemble.}
Given a data set  $\mathbf{X}=[\mathbf{x}_1,\mathbf{x}_2, \cdots, \mathbf{x}_n]\in\mathbb{R}^{d\times n}$ of $n$ samples with each sample $\mathbf{x}_i\in\mathbb{R}^{d\times 1}$, and $m$ base clusterings $\mathbf{\Pi}=[\bm{\pi}_1,\bm{\pi}_2,\cdots,\bm{\pi}_m]\in\mathbb{R}^{n\times m}$, where each base clustering $\bm{\pi}_i\in\mathbb{R}^{n\times 1}$ is an $n$-dimensional vector with the $j$-th element $\bm{\pi}_i(j)$ indicating the clustering membership of the $j$-th sample $\mathbf{x}_j$ in  $\bm{\pi}_i$. For clustering ensemble, the cluster 
indicators in different base clusterings are generally different. Fig. 1-A shows a toy example of 6 samples and 3 base clusterings. The objective of clustering ensemble is to combine multiple base clusterings to produce better performance than that of the individual one.


\subsubsection{Prior Art.}
Based on how to use $\bm{\Pi}$, we roughly divide  previous clustering ensemble methods into two categories. The methods in the first category treat $\mathbf{\Pi}$ as a representation of samples and then construct a pairwise affinity matrix $\mathbf{P}\in\mathbb{R}^{n\times n}$ accordingly, which can be generally expressed as
\begin{equation}
\min_{\mathbf{P}}f(\mathbf{\Pi},\mathbf{P})+\lambda\phi(\mathbf{P}),
\end{equation}
where $f(\bm{\Pi}, \mathbf{P})$ is the fidelity term and $\phi(\mathbf{P})$ imposes specific regularization on $\mathbf{P}$. For example, in \cite{gao2016robust}, $f(\cdot,\cdot)$ and $\phi(\cdot)$ denote the Frobenius norm and the nuclear norm, respectively, while those in \cite{zhou2019ensemble} are both the Frobenius norm. 
The second kind of methods first transform each base clustering as a connective matrix (as shown in Fig. 1-B), i.e.,  
\begin{equation}
\mathbf{A}_k(i,j)=\delta(\pi_k(i),\pi_k(j)),
\end{equation}
where $\mathbf{A}_k\in\mathbb{R}^{n\times n}$ is the $k$-th connective matrix constructed from $\bm{\pi}_k$, and 
\begin{equation}
\delta(\pi_k(i),\pi_k(j))=\begin{cases}
1&{\rm if~}\pi_k(i)=\pi_k(j)\\
0&{\rm otherwise}.
\end{cases}
\end{equation} 
And then, the methods in the second category build a co-association matrix $\mathbf{A}\in\mathbb{R}^{n\times n}$ \cite{fred2005combining} according to the connective matrices, i.e., 
\begin{equation}
\mathbf{A}(i,j)=\frac{1}{m}\sum_{k=1}^m\mathbf{A}_k(i,j).
\label{CA}
\end{equation}
As the co-association matrix naturally converts the base clusterings to a pairwise similarity measure, it becomes the cornerstone of clustering ensemble. Recently, many advanced co-association matrix construction methods were proposed to enhance the clustering performance, which can be generally unified in the following formula: 
\begin{equation}
\mathbf{A}(i,j)=\sum_{k=1}^m \bm{\omega}(k) \times \mathbf{A}_k(i,j),
\label{WCA}
\end{equation}
where $\bm{\omega}\in\mathbb{R}^{m\times 1}$ is the weight vector constructed with different strategies. 
For example, \cite{zhou2020self} used a self-paced learning strategy to construct $\bm{\omega}$.
\cite{huang2017locally} considered the uncertainties of the base clustering, and proposed a locally-weighted weight vector.  \cite{huang2018enhanced} used the cluster-wise similarities to construct the weight vector.


%
%
%

\section{Proposed Method}
As shown in Eq. \eqref{WCA}, the previous methods construct a co-association matrix as the linear combination of connective matrices, and thus are vulnerable to some poor base clusterings. To this end, we propose a novel low-rank tensor approximation based method to refine the initial co-association matrix from a global perspective.  
\subsection{Problem Formulation} To refine the co-association matrix, we first construct a coherent-link matrix (as shown in Fig. 1-D), which inspects whether two samples are clustered to the same category under all the base clusterings. It is worth pointing out that the elements of the coherent-link matrix are highly  reliable information we could infer from the base clusteirngs. Specifically,  we could directly get the coherent-link matrix $\mathbf{M}\in\mathbb{R}^{n\times n}$ from the co-association matrix in Eq. \eqref{CA}, i.e,  
%
\begin{equation}
\mathbf{M}(i,j)=\begin{cases}
1&{\rm if~}\mathbf{A}(i,j)=1\\
0&{\rm otherwise.}
\end{cases}
\label{CLM}
\end{equation}

We then stack the coherent-link matrix and the co-association matrix to form a $3$-D tensor $\bm{\mathcal{P}}\in\mathbb{R}^{n\times n \times 2}$, with $\bm{\mathcal{P}}(:,:,1)=\mathbf{M}$, and $\bm{\mathcal{P}}(:,:2)=\mathbf{A}$. 
As the elements of both the coherent-link matrix and the co-association matrix express the pairwise similarity between samples, ideally, the formed tensor should be low-rank. Moreover, the non-one  elements of $\mathbf{M}$ are limited but express the highly reliable similarity between samples, and we thus try to complement the zero elements with reference to the non-zero ones and the co-association matrix.
On the contrary, the elements of the co-association matrix is dense but with many error connections, and we try to refine it by removing the incorrect connections which is depicted by $\mathbf{E}\in\mathbb{R}^{n\times n}$, by leveraging the information from the coherent-link matrix. In addition,  the elements of $\bm{\mathcal{P}}$ should be bounded in $[0,1]$, and each frontal slice of $\bm{\mathcal{P}}$ should be symmetric.  
Taking all the above analyses into account, the proposed method is mathematically formulated as a constrained optimization problem, written as 
\begin{equation}
\begin{split}
&\min_{\bm{\mathcal{P}},\mathbf{E}}\|\bm{\mathcal{P}}\|_{\star}+\lambda\|\mathbf{E}\|_F^2\\
&{\rm s.t.}~\bm{\mathcal{P}}(i,j,1)=\mathbf{M}(i,j),~{\rm if}~\mathbf{M}(i,j)=1,\\
&~~~~~~\bm{\mathcal{P}}(:,:,1)=\bm{\mathcal{P}}(:,:,1)^\mathsf{T},0\leq\bm{\mathcal{P}}(i,j,1)\leq1,\forall i,j,\\
&~~~~~~\bm{\mathcal{P}}(:,:,2)+\mathbf{E}=\mathbf{A},\\
&~~~~~~\bm{\mathcal{P}}(:,:,2)=\bm{\mathcal{P}}(:,:,2)^\mathsf{T},0\leq\bm{\mathcal{P}}(i,j,2)\leq1,\forall i,j,
\end{split}
\label{model}
\end{equation}
where $\lambda>0$ is the  coefficient to balance  the error matrix, and a Frobenius norm is imposed on $\mathbf{E}$ to avoid trivial solution, i.e., $\bm{\mathcal{P}}(:,:,2)=0$.  
By optimizing Eq. \eqref{model}, it is expected that the limited but highly reliable information in $\mathbf{M}$ could be  propagated to the co-association matrix, while  the coherent-link matrix is complemented according to the information from the co-association matrix at the same time.

After solving the problem in Eq. \eqref{model}, we can obtain a refined co-association matrix $\bm{\mathcal{P}}^*(:,:,2)$ with $\bm{\mathbf{P}}^*$ being the optimized solution. Then, one can apply any clustering methods based on pairwise similarity on $\bm{\mathcal{P}}^*(:,:,2)$ to generate the final clustering result. In this paper, we investigate two popular clustering methods, i.e.,  spectral clustering \cite{ng2002spectral} and agglomerative hierarchical clustering \cite{fred2005combining}. 
\subsection{Numerical Solution}
We propose an optimization method to solve Eq. \eqref{model}, based on the inexact Augmented Lagrangian method \cite{8253493}. Specifically, we first introduce two auxiliary matrices $\mathbf{B}, \mathbf{C}\in\mathbb{R}^{n\times n}$ to deal with the bounded and symmetric constraints on $\bm{\mathcal{P}}(:,:,1)$ and $\bm{\mathcal{P}}(:,:,2)$, respectively, and  Eq. \eqref{model} can be equivalently rewritten as 
\begin{equation}
\begin{split}
&\argmin_{\bm{\mathcal{P}},\mathbf{E},\mathbf{B},\mathbf{C}}\|\bm{\mathcal{P}}\|_{\star}+\lambda\|\mathbf{E}\|_F^2\\
&{\rm s.t.}~\mathbf{B}(i,j)=\mathbf{M}(i,j),~{\rm if}~\mathbf{M}(i,j)=1,~\mathbf{B}=\mathbf{B}^\mathsf{T},\\
&~~~~~~0\leq\mathbf{B}(i,j)\leq1,\forall i,j,~\mathbf{B}=\bm{\mathcal{P}}(:,:,1),\\
&~~~~~~\bm{\mathcal{P}}(:,:,2)+\mathbf{E}=\mathbf{A},~\mathbf{C}=\bm{\mathcal{P}}(:,:,2),\\
&~~~~~~\mathbf{C}=\mathbf{C}^\mathsf{T},~0\leq\mathbf{C}(i,j)\leq1,\forall i,j.
\end{split}
\label{BC}
\end{equation}
To handle the  equality constraints, we introduce three  
Lagrange multipliers $\bm{\Lambda}_1, \bm{\Lambda}_2$ and $\bm{\Lambda}_3\in\mathbb{R}^{n\times n}$, and the augmented Lagrangian form of Eq. \eqref{BC} becomes
\begin{align}
&\argmin_{\bm{\mathcal{P}},\mathbf{E},\mathbf{B},\mathbf{C}}\|\bm{\mathcal{P}}\|_{\star}+\lambda\|\mathbf{E}\|_F^2+\frac{\mu}{2}\left\|\bm{\mathcal{P}}(:,:,2)+\mathbf{E}-\mathbf{A}+\frac{\mathbf{\Lambda}_2}{\mu}\right\|_F^2\nonumber\\
&+\frac{\mu}{2}\left\|\bm{\mathcal{P}}(:,:,1)-\mathbf{B}+\frac{\mathbf{\Lambda}_1}{\mu}\right\|_F^2+\frac{\mu}{2}\left\|\bm{\mathcal{P}}(:,:,2)-\mathbf{C}+\frac{\mathbf{\Lambda}_3}{\mu}\right\|_F^2\nonumber\\
&{\rm s.t.}~\mathbf{B}(i,j)=\mathbf{M}(i,j),~{\rm if}~\mathbf{M}(i,j)=1,0\leq\mathbf{B}(i,j)\leq1,\forall i,j, \nonumber\\
&~~~~~~~\mathbf{B}=\mathbf{B}^\mathsf{T},  \mathbf{C}=\mathbf{C}^\mathsf{T},~0\leq\mathbf{C}(i,j)\leq1,\forall i,j, \label{ALM}
\end{align}
where $\mu>0$ is the penalty coefficient. Then Eq. \eqref{ALM} can be optimized by solving the following four subproblems iteratively and alternately, i.e., only one variable is updated with the remaining ones fixed at each time. 

\subsubsection{The $\bm{\mathcal{P}}$ subproblem.} Removing the irrelevant terms, Eq. \eqref{ALM} with respect to $\bm{\mathcal{P}}$ is written as 
%
\begin{equation}
\begin{split}
&\argmin_{\bm{\mathcal{P}}}\frac{1}{\mu}\|\bm{\mathcal{P}}\|_{\star}+\frac{1}{2}\left\|\bm{\mathcal{P}}-\bm{\mathcal{T}}\right\|_F^2,\\
\end{split}
\label{P-sub1}
\end{equation}
where 
\begin{equation}
\begin{cases}
\bm{\mathcal{T}}(:,:,1)=\mathbf{B}-\frac{\mathbf{\Lambda}_1}{\mu}\\
\bm{\mathcal{T}}(:,:,2)=\frac{1}{2}\left(\mathbf{A}+\mathbf{C}-\mathbf{E}-\frac{\mathbf{\Lambda}_2+\mathbf{\Lambda}_3}{\mu}\right).
\end{cases}
\end{equation}
According to \cite{zhang2014novel}, Eq. \eqref{P-sub1} has a closed-form solution with the soft-thresholding operator of the tensor singular values.  Moreover, according to Algorithm 1, t-SVD computes FFT and SVD on the frontal slices of the input $3$-D tensor $\bm{\mathcal{T}}(:,:,i)$ and its FFT version $\bm{\mathcal{T}}_f(:,:.i)$, respectively, which mainly emphasizes the low-rankness of the frontal slices. Differently, we aim to take advantage of the correction between the original co-association matrix and the coherent-link matrix. Therefore, we perform FFT and SVD on the lateral slices of the tensors $\bm{\mathcal{T}}(:,i,:)$, and  $\bm{\mathcal{T}}_f(:,i,:)$, respectively, to get the t-SVD representation. 
\subsubsection{The $\mathbf{E}$ subproblem.} Without the irrelevant terms, the $\mathbf{E}$ subproblem becomes:
\begin{equation}
\begin{split}
&\min_{\mathbf{E}}\lambda\|\mathbf{E}\|_F^2+\frac{\mu}{2}\left\|\bm{\mathcal{P}}(:,:,2)+\mathbf{E}-\mathbf{A}+\frac{\mathbf{\Lambda}_2}{\mu}\right\|_F^2.\\
\end{split}
\label{E-sub}
\end{equation}
Since Eq. \eqref{E-sub} is quadratic function of $\mathbf{E}$, we can get its global minimum by setting the  derivative of it to $0$, i.e., 
\begin{equation}
\mathbf{E}=\frac{\mu\mathbf{A}-\mathbf{\Lambda}_2-\mu\bm{\mathcal{P}}(:,:,2)}{2\lambda+\mu}.
\label{E-sol}
\end{equation}

\subsubsection{The $\mathbf{B}$ subproblem.} The $\mathbf{B}$ subproblem is written as 
\begin{equation}
\begin{split}
&\min_{\mathbf{B}}\frac{\mu}{2}\left\|\mathbf{B}-\left(\bm{\mathcal{P}}(:,:,1)+\frac{\mathbf{\Lambda}_1}{\mu}\right)\right\|_F^2\\
&{\rm s.t.}~\mathbf{B}(i,j)=\mathbf{M}(i,j),~{\rm if}~\mathbf{M}(i,j)=1,\\
&~~~~~~~\mathbf{B}=\mathbf{B}^\mathsf{T},~0\leq\mathbf{B}(i,j)\leq1,\forall i,j,\\
\end{split}
\label{B-sub}
\end{equation}
which is a symmetric and bounded constrained least squares problem, and has an optimal solution  in element-wise \cite{9154586}, i.e., 
\begin{equation}
\mathbf{B}(i,j)=\begin{cases}
\mathbf{M}(i,j)&{\rm if~}\mathbf{M}(i,j)=1,\\
0&{\rm if~}\mathbf{T}_1(i,j)\leq0~\&~ \mathbf{M}(i,j)\neq 1,\\
1&{\rm if~}\mathbf{T}_1(i,j)\geq1~\&~ \mathbf{M}(i,j)\neq 1,\\
\mathbf{T}_1(i,j)&{\rm if~}0\leq\mathbf{T}_1(i,j)\leq1~\&~ \mathbf{M}(i,j)\neq 1,\\
\end{cases}
\label{B-sol}
\end{equation}
where 
\begin{equation}
\mathbf{T}_1=\frac{1}{2}\left(\bm{\mathcal{P}}(:,:,1)+\bm{\mathcal{P}}(:,:,1)^\mathsf{T}+\frac{\mathbf{\Lambda}_1+\mathbf{\Lambda}^\mathsf{T}_1}{\mu}\right).
\end{equation}
\subsubsection{The $\mathbf{C}$ subproblem.} The $\mathbf{C}$ subproblem  is identical to the $\mathbf{B}$ subproblem without a set of  element-wise equality constraints, which is written as
\begin{equation}
\begin{split}
&\min_{\mathbf{C}}\frac{\mu}{2}\left\|\mathbf{C}-\left(\bm{\mathcal{P}}(:,:,2)+\frac{\mathbf{\Lambda}_3}{\mu}\right)\right\|_F^2\\
&{\rm s.t.}~\mathbf{C}=\mathbf{C}^\mathsf{T},~0\leq\mathbf{C}(i,j)\leq1,\forall i,j,
\end{split}
\label{C-sub}
\end{equation}
and the optimal solution of it is 
\begin{equation}
\mathbf{C}(i,j)=\begin{cases}
\mathbf{T}_2(i,j)&{\rm~if~}0\leq\mathbf{T}_2(i,j)\leq1,\\
0&{\rm~if~}\mathbf{T}_2(i,j)\leq0,\\
1&{\rm~if~}\mathbf{T}_2(i,j)\geq1,\\
\end{cases}
\label{C-sol}
\end{equation}
where 
\begin{equation}
\mathbf{T}_2=\frac{1}{2}\left(\bm{\mathcal{P}}(:,:,2)+\bm{\mathcal{P}}(:,:,2)^\mathsf{T}+\frac{\mathbf{\Lambda}_3+\mathbf{\Lambda}^\mathsf{T}_3}{\mu}\right).
\end{equation}

\subsubsection{Update $\mathbf{\Lambda}_1$, $\mathbf{\Lambda}_2$, $\mathbf{\Lambda}_3$ and $\mu$} The Lagrange multipliers and $\mu$ are updated by 
\begin{equation}
\begin{cases}
\mathbf{\Lambda}_1=\mathbf{\Lambda}_1+\mu(\bm{\mathcal{P}}(:,:,1)-\mathbf{B})\\
\mathbf{\Lambda}_2=\mathbf{\Lambda}_2+\mu(\bm{\mathcal{P}}(:,:,2)+\mathbf{E}-\mathbf{A})\\
\mathbf{\Lambda}_3=\mathbf{\Lambda}_3+\mu(\bm{\mathcal{P}}(:,:,2)-\mathbf{C})\\
\mu={\rm min}(1.1\mu,\mu_{\rm max}),\\
\end{cases}
\label{mu-sol}
\end{equation}
where $\mu$ is initialized to $0.0001$ \cite{liu2019imbalance}, and $\mu_{\rm max}$ is the upper-bound for $\mu$. The overall numerical solution is summarized in Algorithm \ref{TLA}, where the stopping conditions is ${\rm max}(\|\mathbf{B}-\bm{\mathcal{P}}(:,:,1)\|_\infty, \|\mathbf{C}-\bm{\mathcal{P}}(:,:,2)\|_\infty, \|\mathbf{A}-\mathbf{E}-\bm{\mathcal{P}}(:,:,2)\|_\infty )<10^{-8}$ with $\|\cdot\|_\infty$ being the maximum of the absolute values of a matrix. 
\begin{algorithm}[!t]
	\caption{Numerical solution to Eq. \eqref{model}}
	\begin{algorithmic}[1]
		\renewcommand{\algorithmicrequire}{\textbf{Input:}}
		\renewcommand{\algorithmicensure}{\textbf{Initialize:}}
		\REQUIRE Base clusterings matrix $\mathbf{\Pi}$;
		\ENSURE $\bm{\mathcal{P}}=0$, $\mathbf{E}=0$, $\mathbf{B}=0$, $\mathbf{C}=0$, and $\mu_{\rm max}=10^8$;
		\STATE Construct the co-association matrix $\mathbf{A}$ by Eq. \eqref{CA};
		\STATE Construct the coherent-link matrix $\mathbf{M}$ by Eq. \eqref{CLM};
		\WHILE{not converged }
		\STATE Update $\bm{\mathcal{P}}$ by solving Eq. \eqref{P-sub1};
		\STATE Update $\mathbf{E}$ by Eq. \eqref{E-sol};
		\STATE Update $\mathbf{B}$ by Eq. \eqref{B-sol};
		\STATE Update $\mathbf{C}$ by Eq. \eqref{C-sol};
		\STATE Update $\mathbf{\Lambda}_1$, $\mathbf{\Lambda}_2$, $\mathbf{\Lambda}_3$ and $\mu$ by Eq. \eqref{mu-sol};
		\STATE Check the convergence conditions;
		%
		\ENDWHILE
	\end{algorithmic}
	\textbf{Output:} $\bm{\mathcal{P}}(:,:,2)$ as the refined co-association matrix. 
	\label{TLA}
\end{algorithm}

\section{Experiment}
We conducted extensive experiments to evaluate the proposed model. 
To reproduce the results, we made the code publicly
available at https://github.com/jyh-learning/TensorClusteringEnsemble.

\subsubsection{Data Sets.}
Following recent clustering ensemble papers \cite{huang2017locally, huang2015robust, zhou2019ensemble}, we adopted $7$ commonly used data sets, i.e., BinAlpha, Multiple features (MF), MNIST, Semeion, CalTech, Texture and ISOLET. 
Following \cite{huang2017locally}, we randomly selected 5000 samples from MNIST and used the subset in the experiments, and for CalTech, we used 20 representative categories out of $101$ categories and denoted it as CalTech20.   
\subsubsection{Generation of Base Clusterings.}
Following \cite{huang2017locally}, we first generated a pool of $100$ candidate base clusterings for all the data sets by applying the the K-means algorithm with the value of K randomly varying in  the range of $[2,\sqrt{n}]$, where $n$ is the number of input data samples.  
\subsubsection{Methods under Comparison.}
We compared the proposed model with $12$ state-of-the-art clustering  ensemble methods, including PTA-AL, PTA-CL, PTA-SL and PTGP \cite{ huang2015robust}, LWSC, LWEA and LWGP \cite{huang2017locally}, ECPCS-HC and ECPCS-MC \cite{huang2018enhanced}, DREC \cite{zhou2019ensemble}, SPCE \cite{zhou2020self},  and SEC \cite{7811216}. 
The codes of all the compared methods are provided by the authors.
Ours-EA and Ours-SC denote the proposed model equipped with  agglomerative hierarchical 
clustering  and spectral clustering, respectively, to generate the final clustering result.  

\subsubsection{Evaluation Metrics.}
We adopted $7$ commonly used metrics to evaluate clustering performance, i.e., clustering accuracy (ACC), normalized mutual information (NMI), purity, adjust rand index (ARI), F$1$-score, precision, and recall. For all the metrics, a larger value indicates better clustering performance, and the values of all the metrics are up-bounded by $1$. The detailed definitions of those metrics can be found in \cite{8502831,9072553}. 
\subsubsection{Experiment Settings.}
For each data set, we randomly selected $10$ base clusterings from the candidate base clustering pool, and performed different  clustering ensemble methods on the selected base clusteirngs. To reduce the influence of the selected  base clusterings, we repeated the random selection $20$ times, and reported the average performance over the $20$ repetitions. 
For the compared methods, we set the  hyper-parameters according to their original papers. If there are no  suggested values, we exhaustively searched the hyper-parameters, and used the ones producing the best performance. 
The proposed model only contains one hyper-parameter $\lambda$, which was set to $0.002$ for all the data sets. 
\begin{table*}[tph]
	\caption{Clustering Performance on BinAlpha (\# samples: $1404$, dimension:  $320$, \# clusters:  $36$) }\smallskip
	\vspace{-0.5\baselineskip}
	\centering
	\resizebox{2.05\columnwidth}{!}{
		\smallskip\begin{tabular}{l c c c c c c c c c c c c c c}
			\hline
			\textbf{BinAlpha}	&	PTA-AL	&	PTA-CL	&	PTA-SL	&	PTGP	&	LWSC	&	LWEA	&	LWGP	&	ECPCS-HC	&	ECPCS-MC	&	DREC	&	SPCE	&	SEC	&	Ours-EA	&	Ours-SC	\\ \hline
			ACC	&$	0.430 	$&$	0.429 	$&$	0.186 	$&$	0.429 	$&$	0.424 	$&$	0.403 	$&$	0.431 	$&$	0.375 	$&$	0.454 	$&$	0.375 	$&$	0.298	$&$	0.443 	$&$\underline{	0.712 	}$&$\mathbf{	0.858 	}$\\
			NMI	&$	0.574 	$&$	0.577 	$&$	0.300 	$&$	0.574 	$&$	0.570 	$&$	0.553 	$&$	0.575 	$&$	0.537 	$&$	0.592 	$&$	0.518 	$&$	0.541 	$&$	0.585 	$&$\underline{	0.824 	}$&$\mathbf{	0.916 	}$\\
			Purity	&$	0.447 	$&$	0.451 	$&$	0.197 	$&$	0.446 	$&$	0.444 	$&$	0.413 	$&$	0.457 	$&$	0.383 	$&$	0.478 	$&$	0.396 	$&$	0.285	$&$	0.470 	$&$\underline{	0.718 	}$&$\mathbf{	0.876 	}$\\
			ARI	&$	0.292 	$&$	0.291 	$&$	0.081 	$&$	0.291 	$&$	0.284 	$&$	0.289 	$&$	0.287 	$&$	0.269 	$&$	0.300 	$&$	0.248 	$&$	0.227	$&$	0.291 	$&$\underline{	0.643 	}$&$\mathbf{	0.817 	}$\\
			F1-score	&$	0.314 	$&$	0.312 	$&$	0.126 	$&$	0.313 	$&$	0.306 	$&$	0.313 	$&$	0.308 	$&$	0.295 	$&$	0.320 	$&$	0.271 	$&$	0.302 	$&$	0.311 	$&$\underline{	0.654 	}$&$\mathbf{	0.822 	}$\\
			Precision	&$	0.275 	$&$	0.276 	$&$	0.071 	$&$	0.277 	$&$	0.272 	$&$	0.248 	$&$	0.277 	$&$	0.220 	$&$	0.305 	$&$	0.238 	$&$	0.294 	$&$	0.296 	$&$\underline{	0.559 	}$&$\mathbf{	0.801 	}$\\
			Recall	&$	0.366 	$&$	0.361 	$&$	0.635 	$&$	0.361 	$&$	0.349 	$&$	0.426 	$&$	0.348 	$&$	0.451 	$&$	0.337 	$&$	0.323 	$&$	0.314 	$&$	0.327 	$&$\underline{	0.791 	}$&$\mathbf{	0.845 	}$\\
			\hline
		\end{tabular}
	}
	\begin{tablenotes}
		\item[*] \tiny{The highest value in each row is bolded, and  the second highest one is underlined.}\
	\end{tablenotes}
	\label{table-BA}
\end{table*}
\begin{table*}[tph]
	\caption{Clustering Performance on MF (\# samples: 2000,  dimension:  649, \# clusters:  10)}\smallskip
	\vspace{-0.5\baselineskip}
	\centering
	\resizebox{2.05\columnwidth}{!}{
		\smallskip\begin{tabular}{l c c c c c c c c c c c c c c}
			\hline
			\textbf{MF}	&	PTA-AL	&	PTA-CL	&	PTA-SL	&	PTGP	&	LWSC	&	LWEA	&	LWGP	&	ECPCS-HC	&	ECPCS-MC	&	DREC	&	SPCE	&	SEC	&	Ours-EA	&	Ours-SC	\\ \hline
			ACC	&$	0.647 	$&$	0.606 	$&$	0.507 	$&$	0.648 	$&$	0.671 	$&$	0.609 	$&$	0.649 	$&$	0.589 	$&$	0.652 	$&$	0.362 	$&$	0.581 	$&$	0.592 	$&$\underline{	0.718 	}$&$\mathbf{	0.990 	}$\\
			NMI	&$	0.655 	$&$	0.638 	$&$	0.536 	$&$	0.654 	$&$	0.655 	$&$	0.650 	$&$	0.655 	$&$	0.618 	$&$	0.652 	$&$	0.347 	$&$	0.621 	$&$	0.602 	$&$\underline{	0.790 	}$&$\mathbf{	0.979 	}$\\
			Purity	&$	0.675 	$&$	0.644 	$&$	0.533 	$&$	0.677 	$&$	0.690 	$&$	0.650 	$&$	0.673 	$&$	0.616 	$&$	0.676 	$&$	0.387 	$&$	0.615 	$&$	0.623 	$&$\underline{	0.719 	}$&$\mathbf{	0.990 	}$\\
			ARI	&$	0.523 	$&$	0.500 	$&$	0.371 	$&$	0.523 	$&$	0.533 	$&$	0.514 	$&$	0.530 	$&$	0.481 	$&$	0.526 	$&$	0.257 	$&$	0.459 	$&$	0.472 	$&$\underline{	0.685 	}$&$\mathbf{	0.979 	}$\\
			F1-score	&$	0.574 	$&$	0.554 	$&$	0.457 	$&$	0.575 	$&$	0.583 	$&$	0.567 	$&$	0.582 	$&$	0.541 	$&$	0.576 	$&$	0.370 	$&$	0.527 	$&$	0.528 	$&$\underline{	0.724 	}$&$\mathbf{	0.981 	}$\\
			Precision	&$	0.535 	$&$	0.511 	$&$	0.344 	$&$	0.534 	$&$	0.551 	$&$	0.517 	$&$	0.530 	$&$	0.472 	$&$	0.541 	$&$	0.311 	$&$	0.424 	$&$	0.496 	$&$\underline{	0.586 	}$&$\mathbf{	0.981 	}$\\
			Recall	&$	0.625 	$&$	0.608 	$&$	0.712 	$&$	0.627 	$&$	0.619 	$&$	0.628 	$&$	0.647 	$&$	0.637 	$&$	0.618 	$&$	0.739 	$&$	0.713 	$&$	0.566 	$&$\underline{	0.960 	}$&$\mathbf{	0.981 	}$\\ 
			\hline
		\end{tabular}
	}
	\label{table-MF}
\end{table*}
\begin{table*}[tph]
	\caption{Clustering Performance on MNIST (\# samples: 5000, dimension:  784, \# clusters:  10)}\smallskip
	\vspace{-0.5\baselineskip}
	\centering
	\resizebox{2.05\columnwidth}{!}{
		\smallskip\begin{tabular}{l c c c c c c c c c c c c c c}
			\hline
			\textbf{MNIST}	&	PTA-AL	&	PTA-CL	&	PTA-SL	&	PTGP	&	LWSC	&	LWEA	&	LWGP	&	ECPCS-HC	&	ECPCS-MC	&	DREC	&	SPCE	&	SEC	&	Ours-EA	&	Ours-SC	\\ \hline
			ACC	&$	0.663 	$&$	0.654 	$&$	0.207 	$&$	0.665 	$&$	0.613 	$&$	0.658 	$&$	0.573 	$&$	0.609 	$&$	0.656 	$&$	0.480 	$&$	0.543 	$&$	0.539 	$&$\underline{	0.797 	}$&$\mathbf{	0.977 	}$\\
			NMI	&$	0.618 	$&$	0.610 	$&$	0.133 	$&$	0.622 	$&$	0.612 	$&$	0.635 	$&$	0.594 	$&$	0.608 	$&$	0.635 	$&$	0.434 	$&$	0.482 	$&$	0.521 	$&$\underline{	0.806 	}$&$\mathbf{	0.979 	}$\\
			Purity	&$	0.682 	$&$	0.668 	$&$	0.209 	$&$	0.685 	$&$	0.663 	$&$	0.676 	$&$	0.626 	$&$	0.624 	$&$	0.691 	$&$	0.498 	$&$	0.557 	$&$	0.585 	$&$\underline{	0.798 	}$&$\mathbf{	0.980 	}$\\
			ARI	&$	0.513 	$&$	0.504 	$&$	0.051 	$&$	0.522 	$&$	0.483 	$&$	0.531 	$&$	0.460 	$&$	0.495 	$&$	0.524 	$&$	0.342 	$&$	0.429 	$&$	0.384 	$&$\underline{	0.735 	}$&$\mathbf{	0.969 	}$\\
			F1-score	&$	0.566 	$&$	0.557 	$&$	0.219 	$&$	0.572 	$&$	0.540 	$&$	0.582 	$&$	0.522 	$&$	0.558 	$&$	0.574 	$&$	0.427 	$&$	0.445 	$&$	0.450 	$&$\underline{	0.767 	}$&$\mathbf{	0.972 	}$\\
			Precision	&$	0.520 	$&$	0.523 	$&$	0.124 	$&$	0.541 	$&$	0.490 	$&$	0.536 	$&$	0.459 	$&$	0.448 	$&$	0.543 	$&$	0.373 	$&$	0.316 	$&$	0.420 	$&$\underline{	0.666 	}$&$\mathbf{	0.968 	}$\\
			Recall	&$	0.624 	$&$	0.596 	$&$	0.952 	$&$	0.607 	$&$	0.603 	$&$	0.641 	$&$	0.609 	$&$	0.745 	$&$	0.610 	$&$	0.576 	$&$	0.831 	$&$	0.485 	$&$\underline{	0.918 	}$&$\mathbf{	0.977 	}$\\ \hline
		\end{tabular}
	}
	\label{table-MNIST}
\end{table*}
\begin{table*}[tph]
	\caption{Clustering Performance on Semeion  (\# samples: 1593, dimension:  256, \# clusters:  10)}\smallskip
	\vspace{-0.5\baselineskip}
	\centering
	\resizebox{2.05\columnwidth}{!}{
		\smallskip\begin{tabular}{l c c c c c c c c c c c c c c}\hline
			\textbf{Semeion}	&	PTA-AL	&	PTA-CL	&	PTA-SL	&	PTGP	&	LWSC	&	LWEA	&	LWGP	&	ECPCS-HC	&	ECPCS-MC	&	DREC	&	SPCE	&	SEC	&	Ours-EA	&	Ours-SC	\\ \hline
			ACC	&$	0.688 	$&$	0.700 	$&$	0.425 	$&$	0.692 	$&$	0.682 	$&$	0.739 	$&$	0.620 	$&$	0.638 	$&$	0.679 	$&$	0.450 	$&$	0.571 	$&$	0.594 	$&$\underline{	0.846 	}$&$\mathbf{	0.983 	}$\\
			NMI	&$	0.633 	$&$	0.634 	$&$	0.418 	$&$	0.631 	$&$	0.630 	$&$	0.656 	$&$	0.598 	$&$	0.601 	$&$	0.635 	$&$	0.386 	$&$	0.571 	$&$	0.569 	$&$\underline{	0.831 	}$&$\mathbf{	0.962 	}$\\
			Purity	&$	0.698 	$&$	0.707 	$&$	0.449 	$&$	0.703 	$&$	0.702 	$&$	0.739 	$&$	0.651 	$&$	0.645 	$&$	0.705 	$&$	0.460 	$&$	0.607 	$&$	0.634 	$&$\underline{	0.847 	}$&$\mathbf{	0.983 	}$\\
			ARI	&$	0.507 	$&$	0.510 	$&$	0.248 	$&$	0.507 	$&$	0.507 	$&$	0.540 	$&$	0.465 	$&$	0.480 	$&$	0.508 	$&$	0.290 	$&$	0.401 	$&$	0.418 	$&$\underline{	0.790 	}$&$\mathbf{	0.962 	}$\\
			F1-score	&$	0.561 	$&$	0.563 	$&$	0.360 	$&$	0.560 	$&$	0.559 	$&$	0.588 	$&$	0.525 	$&$	0.540 	$&$	0.560 	$&$	0.391 	$&$	0.477 	$&$	0.481 	$&$\underline{	0.813 	}$&$\mathbf{	0.966 	}$\\
			Precision	&$	0.513 	$&$	0.522 	$&$	0.246 	$&$	0.522 	$&$	0.523 	$&$	0.552 	$&$	0.466 	$&$	0.468 	$&$	0.527 	$&$	0.329 	$&$	0.381 	$&$	0.448 	$&$\underline{	0.748 	}$&$\mathbf{	0.966 	}$\\
			Recall	&$	0.620 	$&$	0.611 	$&$	0.712 	$&$	0.606 	$&$	0.601 	$&$	0.631 	$&$	0.603 	$&$	0.644 	$&$	0.599 	$&$	0.664 	$&$	0.660 	$&$	0.524 	$&$\underline{	0.893 	}$&$\mathbf{	0.966 	}$\\ \hline
		\end{tabular}
	}
	\label{table-Semeion}
\end{table*}
\begin{table*}[tph]
	\caption{Clustering Performance on CalTech20  (\# samples: 2386, dimension:  30,000, \# clusters:  20)}\smallskip
	\vspace{-0.5\baselineskip}
	\centering
	\resizebox{2.05\columnwidth}{!}{
		\smallskip\begin{tabular}{l c c c c c c c c c c c c c c}
			\hline
			\textbf{CalTech20}	&	PTA-AL	&	PTA-CL	&	PTA-SL	&	PTGP	&	LWSC	&	LWEA	&	LWGP	&	ECPCS-HC	&	ECPCS-MC	&	DREC	&	SPCE	&	SEC	&	Ours-EA	&	Ours-SC	\\ \hline
			ACC	&$	0.345 	$&$	0.343 	$&$	{0.421} 	$&$	0.345 	$&$	0.324 	$&$	0.423 	$&$	0.336 	$&$	0.450 	$&$	0.363 	$&$	0.340 	$&$	\underline{0.495} 	$&$	0.297 	$&$\mathbf{	0.726 	}$&${	0.418 	}$\\
			NMI	&$	0.404 	$&$	0.402 	$&$	0.269 	$&$	0.401 	$&$	0.396 	$&$	0.454 	$&$	0.406 	$&$	0.455 	$&$	0.428 	$&$	0.350 	$&$	0.452 	$&$	0.381 	$&$\underline{	0.620 	}$&$\mathbf{	0.621 	}$\\
			Purity	&$	0.641 	$&$	0.639 	$&$	0.520 	$&$	0.637 	$&$	0.642 	$&$	0.665 	$&$	0.646 	$&$	0.645 	$&$	0.660 	$&$	0.590 	$&$	0.664 	$&$	0.633 	$&$\underline{	0.730 	}$&$\mathbf{	0.788 	}$\\
			ARI	&$	0.261 	$&$	0.265 	$&$	0.184 	$&$	0.267 	$&$	0.222 	$&$	{0.359} 	$&$	0.224 	$&$	0.351 	$&$	0.258 	$&$	0.225 	$&$	\underline{0.395} 	$&$	0.202 	$&$\mathbf{	0.785 	}$&${	0.328 	}$\\
			F1-score	&$	0.334 	$&$	0.337 	$&$	0.363 	$&$	0.338 	$&$	0.291 	$&$	0.432 	$&$	0.298 	$&$	0.437 	$&$	0.332 	$&$	0.316 	$&$	\underline{0.457} 	$&$	0.269 	$&$\mathbf{	0.823 	}$&${	0.384 	}$\\
			Precision	&$	0.553 	$&$	0.561 	$&$	0.284 	$&$	0.563 	$&$	0.529 	$&$	0.612 	$&$	0.510 	$&$	0.538 	$&$	0.543 	$&$	0.479 	$&$	0.503 	$&$	0.525 	$&$\mathbf{	0.764 	}$&$\underline{	0.743 	}$\\
			Recall	&$	0.239 	$&$	0.241 	$&$	\underline{0.562} 	$&$	0.243 	$&$	0.201 	$&$	0.335 	$&$	0.211 	$&$	0.373 	$&$	0.239 	$&$	0.253 	$&$	0.449 	$&$	0.181 	$&$\mathbf{	0.898 	}$&${	0.259 	}$\\ \hline
		\end{tabular}
	}
	\label{CalTech}
\end{table*}

\begin{table*}[tph]
	\caption{Clustering Performance on Texture  (\# samples: 5500, dimension:  20, \# clusters:  11)}\smallskip
	\vspace{-0.5\baselineskip}
	\centering
	\resizebox{2.05\columnwidth}{!}{
		\smallskip\begin{tabular}{l c c c c c c  c c c c c c c c}
			\hline
			\textbf{Texture}	&	PTA-AL	&	PTA-CL	&	PTA-SL	&	PTGP	&	LWSC	&	LWEA	&	LWGP	&	ECPCS-HC	&	ECPCS-MC	&	DREC	&	SPCE	&	SEC	&	Ours-EA	&	Ours-SC	\\ \hline
			ACC	&$	0.741 	$&$	0.714 	$&$	0.410 	$&$	0.732 	$&$	0.719 	$&$	0.793 	$&$	0.686 	$&$	0.675 	$&$	0.675 	$&$	0.416 	$&$	0.634 
			$&$	0.614 	$&$\underline{	0.863 	}$&$\mathbf{	0.993 	}$\\
			NMI	&$	0.742 	$&$	0.721 	$&$	0.438 	$&$	0.731 	$&$	0.742 	$&$	0.782 	$&$	0.739 	$&$	0.703 	$&$	0.718 	$&$	0.419 	$&$	0.693 
			$&$	0.638 	$&$\underline{	0.868 	}$&$\mathbf{	0.995 	}$\\
			Purity	&$	0.751 	$&$	0.729 	$&$	0.427 	$&$	0.746 	$&$	0.744 	$&$	0.798 	$&$	0.728 	$&$	0.685 	$&$	0.698 	$&$	0.441 	$&$	0.658 
			$&$	0.647 	$&$\underline{	0.864 	}$&$\mathbf{	0.995 	}$\\
			ARI	&$	0.628 	$&$	0.600 	$&$	0.237 	$&$	0.619 	$&$	0.628 	$&$	0.696 	$&$	0.609 	$&$	0.569 	$&$	0.585 	$&$	0.298 	$&$	0.534 
			$&$	0.486 	$&$\underline{	0.816 	}$&$\mathbf{	0.993 	}$\\
			F1-score	&$	0.664 	$&$	0.639 	$&$	0.350 	$&$	0.656 	$&$	0.663 	$&$	0.724 	$&$	0.648 	$&$	0.614 	$&$	0.626 	$&$	0.397 	$&$	0.590 
			$&$	0.537 	$&$\underline{	0.834 	}$&$\mathbf{	0.993 	}$\\
			Precision	&$	0.624 	$&$	0.598 	$&$	0.228 	$&$	0.627 	$&$	0.631 	$&$	0.700 	$&$	0.592 	$&$	0.543 	$&$	0.582 	$&$	0.337 	$&$0.465 
			$&$	0.500 	$&$\underline{	0.780 	}$&$\mathbf{	0.991 	}$\\
			Recall	&$	0.712 	$&$	0.690 	$&$	\underline{0.897} 	$&$	0.689 	$&$	0.700 	$&$	0.752 	$&$	0.719 	$&$	0.710 	$&$	0.677 	$&$	0.752 	$&$	0.827 
			$&$	0.586 	$&${	0.895 	}$&$\mathbf{	0.996 	}$\\ \hline
			
		\end{tabular}
	}
	\label{Texture}
\end{table*}

\begin{table*}[tph]
	\caption{Clustering Performance on ISOLET  (\# samples: 7791, dimension:  617, \# clusters:  26)}\smallskip
	\vspace{-0.5\baselineskip}
	\centering
	\resizebox{2.05\columnwidth}{!}{
		\smallskip\begin{tabular}{l c c c c c c c c c c c c c c}
			\hline
			\textbf{ISOLET}	&	PTA-AL	&	PTA-CL	&	PTA-SL	&	PTGP	&	LWSC	&	LWEA	&	LWGP	&	ECPCS-HC	&	ECPCS-MC	&	DREC	&	SPCE	&	SEC	&	Ours-EA	&	Ours-SC	\\ \hline
			ACC	&$	0.551 	$&$	0.540 	$&$	0.394 	$&$	0.539 	$&$	0.556 	$&$	0.578 	$&$	0.527 	$&$	0.451 	$&$	\underline{0.581} 	$&$	0.324 	$&$	0.574	$&$	0.554 	$&${	0.575 	}$&$\mathbf{	0.675 	}$\\
			NMI	&$	0.722 	$&$	0.718 	$&$	0.587 	$&$	0.715 	$&$	0.721 	$&$	0.743 	$&$	0.710 	$&$	0.667 	$&$	0.743 	$&$	0.413 	$&$\underline{	0.818} 
			$&$	0.719 	$&${	0.752 	}$&$\mathbf{	0.831 	}$\\
			Purity	&$	0.580 	$&$	0.572 	$&$	0.407 	$&$	0.567 	$&$	0.594 	$&$	0.605 	$&$	0.564 	$&$	0.467 	$&$	\underline{0.619} 	$&$	0.350 	$&$	0.301 
			$&$	0.590 	$&${	0.583 	}$&$\mathbf{	0.707 	}$\\
			ARI	&$	0.507 	$&$	0.498 	$&$	0.316 	$&$	0.495 	$&$	0.485 	$&$	0.552 	$&$	0.472 	$&$	0.449 	$&$	0.516 	$&$	0.251 	$&$	0.367	$&$	0.483 	$&$\underline{	0.563 	}$&$\mathbf{	0.639 	}$\\
			F1-score	&$	0.529 	$&$	0.520 	$&$	0.356 	$&$	0.517 	$&$	0.506 	$&$	0.571 	$&$	0.495 	$&$	0.477 	$&$	0.536 	$&$	0.303 	$&$	0.384 
			$&$	0.505 	$&$\underline{	0.584 	}$&$\mathbf{	0.654 	}$\\
			Precision	&$	0.479 	$&$	0.467 	$&$	0.237 	$&$	0.462 	$&$	0.458 	$&$	0.511 	$&$	0.437 	$&$	0.352 	$&$	{0.496} 	$&$	0.253 	$&$	\underline{0.584} 
			$&$	0.463 	$&${	0.481 	}$&$\mathbf{	0.625 	}$\\
			Recall	&$	0.593 	$&$	0.591 	$&$	\mathbf{0.787} 	$&$	0.590 	$&$	0.568 	$&$	0.648 	$&$	0.573 	$&$	0.748 	$&$	0.584 	$&$	0.690 	$&$	0.327	$&$	{0.555} 	$&$\underline{	0.752 	}$&${	0.685 	}$\\ \hline
			
		\end{tabular}
	}
	\label{ISOLET}
\end{table*}

\subsection{Analysis of the Clustering Performance}

\begin{figure}[!t]
	\centering
	\centerline{\epsfig{figure=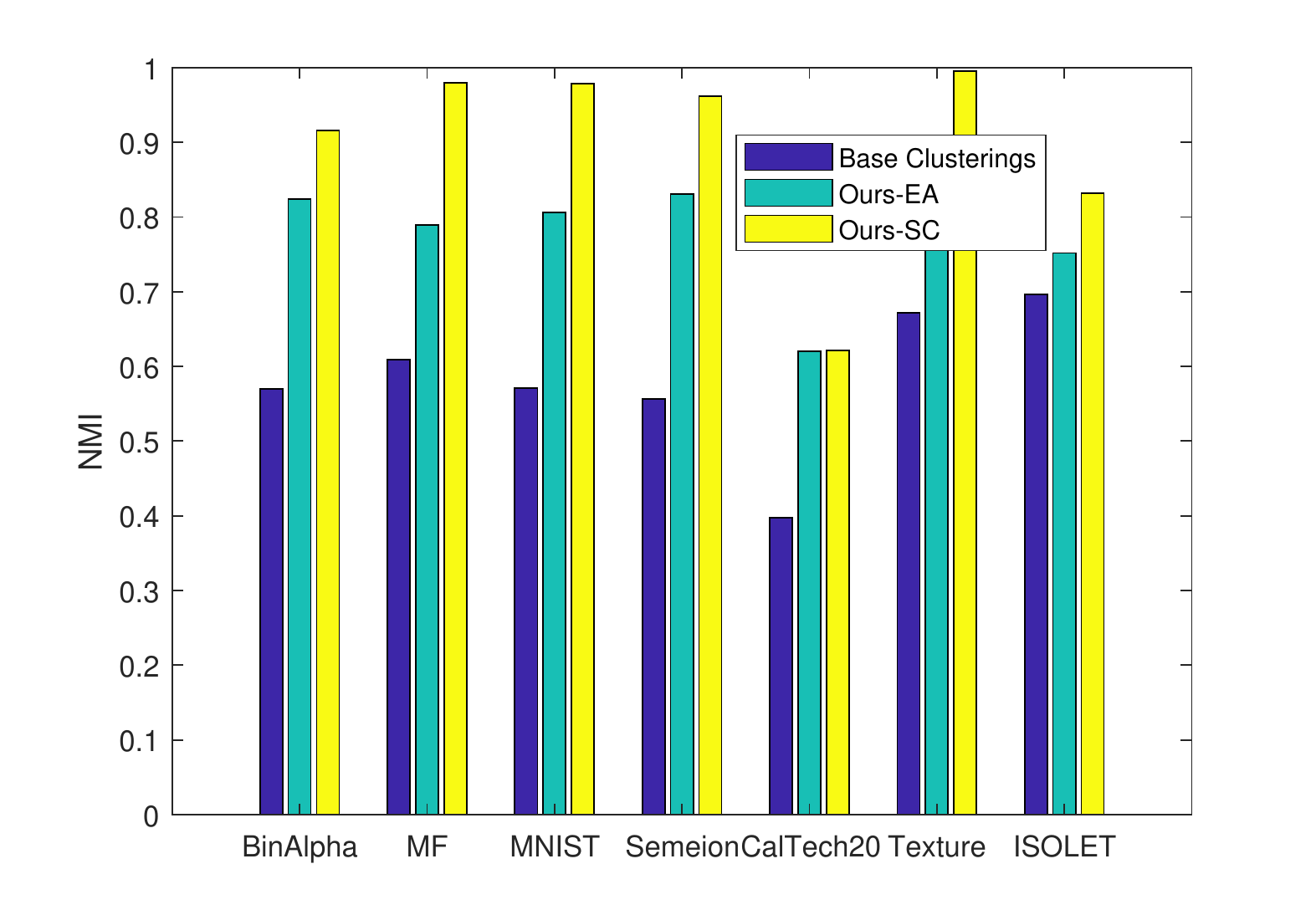,width=7.8cm}}
	\caption{The NMI of our methods against the average NMI of the base clusteings in the candidate base clustering pool. }
	\label{fig:BPnmi}
\end{figure}
\begin{figure}[!t]
	\begin{minipage}[b]{0.49\linewidth}
		\centering
		\centerline{\epsfig{figure=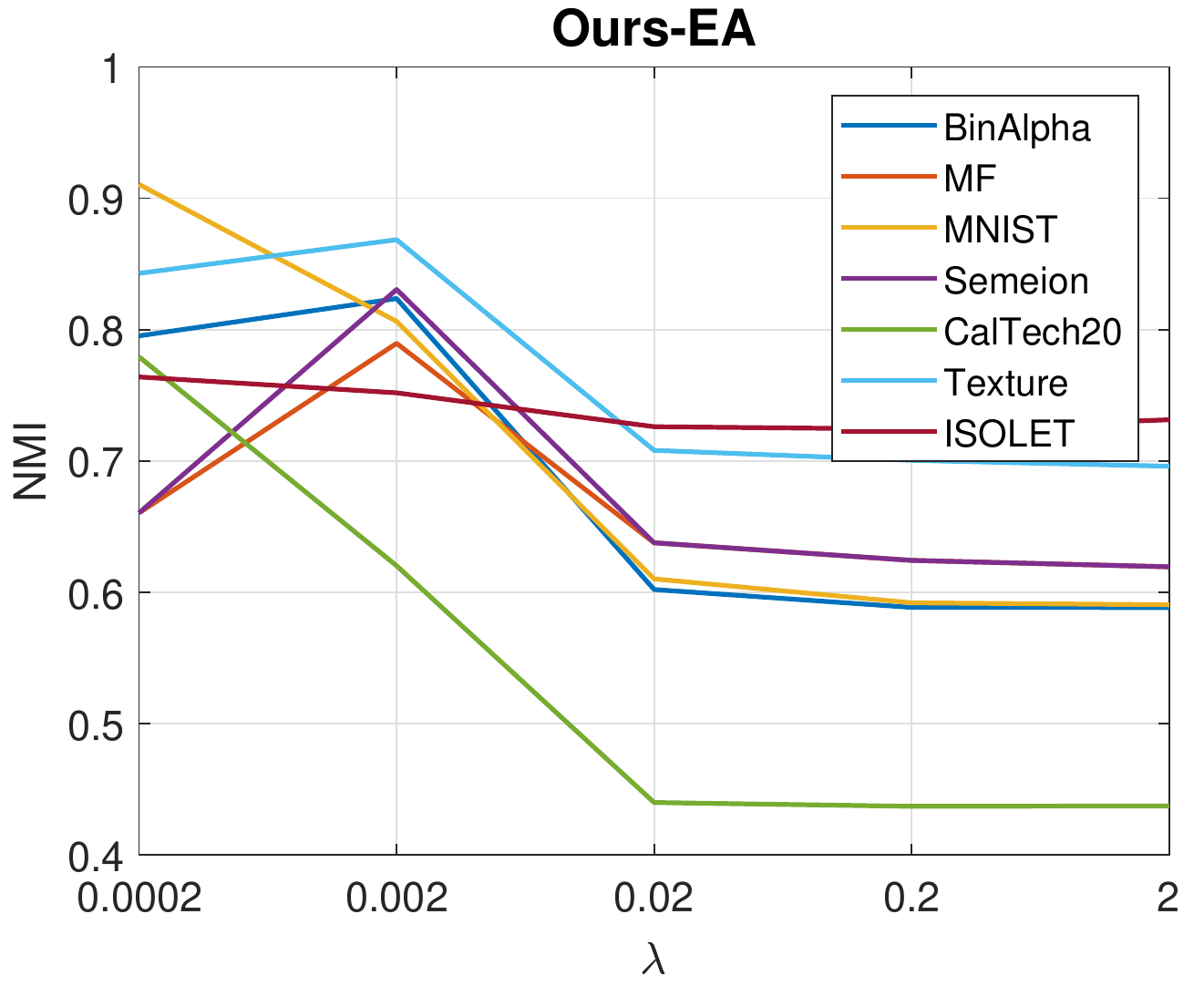,width=4.2cm}}
	\end{minipage}
	\begin{minipage}[b]{0.49\linewidth}
		\centering
		\centerline{\epsfig{figure=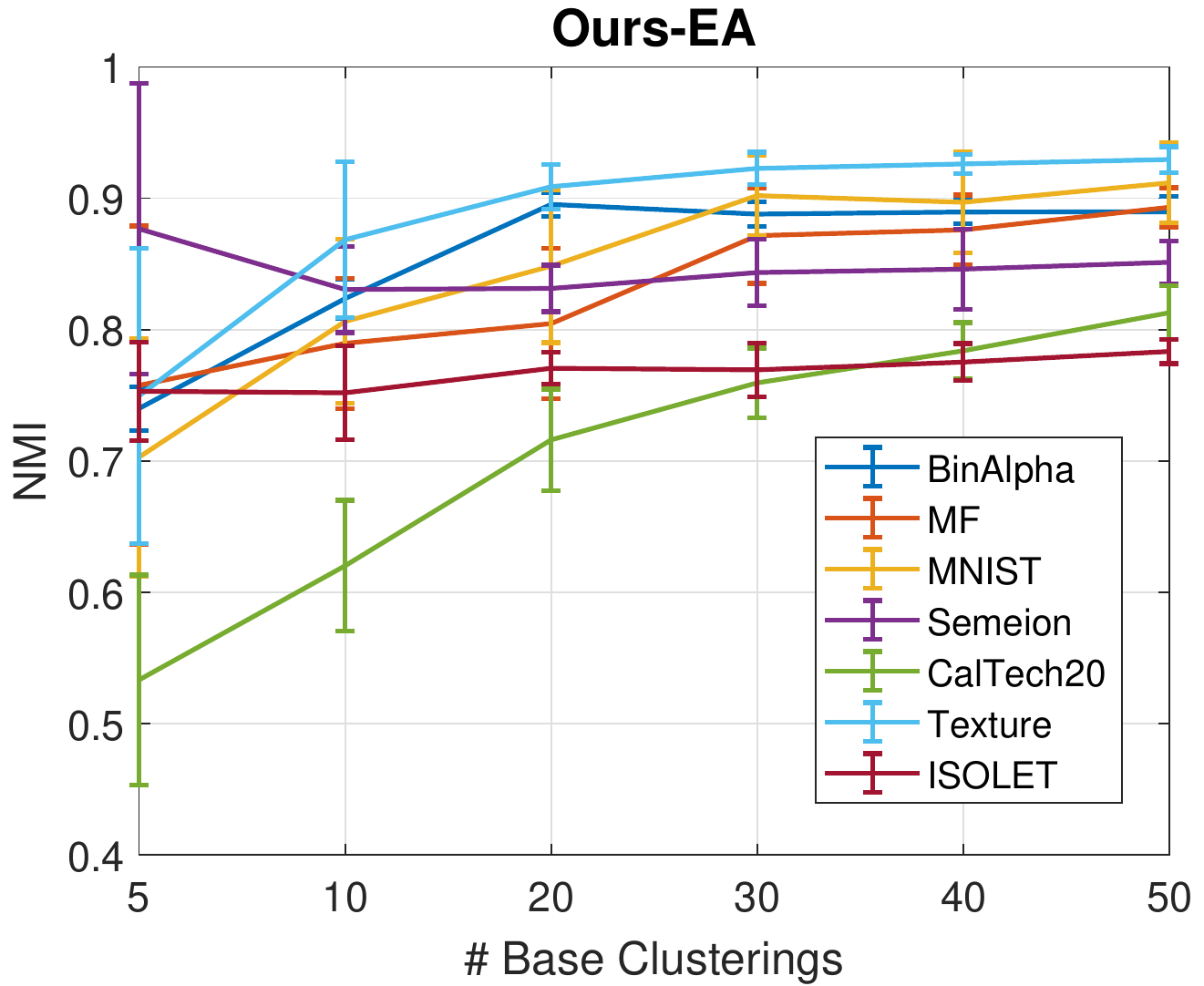,width=4.2cm}}
	\end{minipage}
	\caption{The NMI of our methods against different $\lambda$. }
	\label{fig:lambda}
\end{figure}
\begin{figure}[!t]
	\begin{minipage}[b]{0.49\linewidth}
		\centering
		\centerline{\epsfig{figure=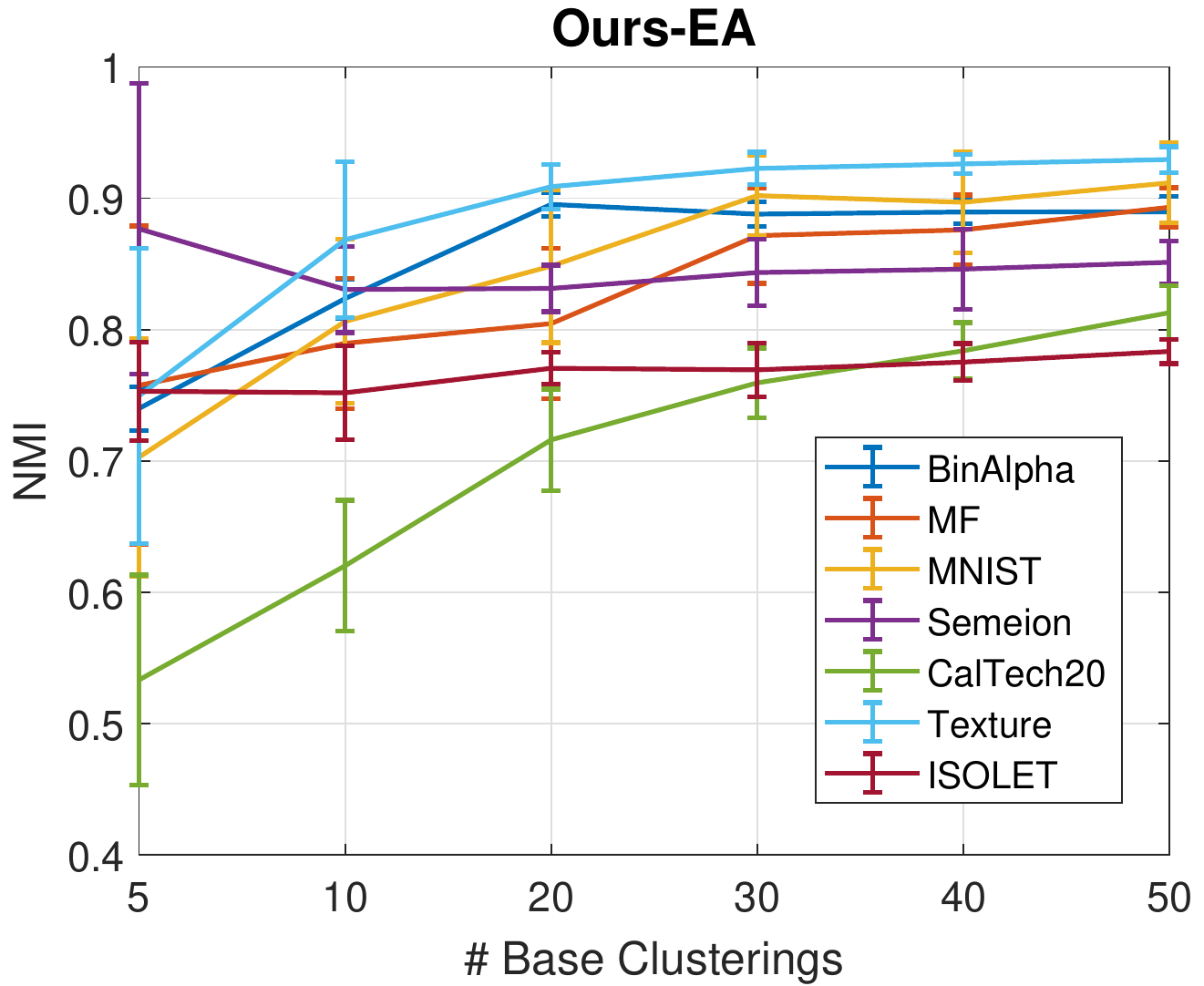,width=4.2cm}}
	\end{minipage}
	\begin{minipage}[b]{0.49\linewidth}
		\centering
		\centerline{\epsfig{figure=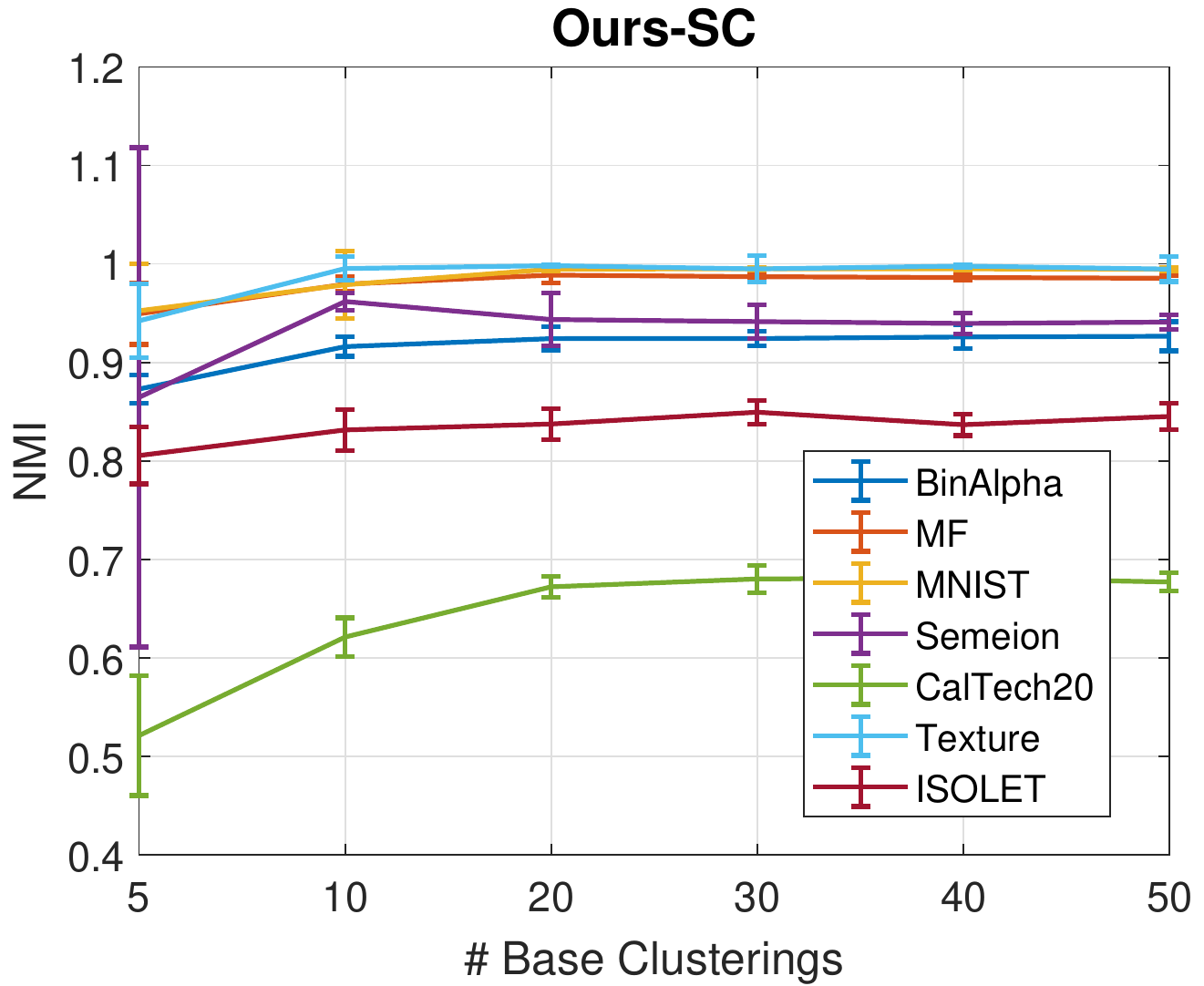,width=4.2cm}}
	\end{minipage}
	\caption{The NMI of our methods  with different numbers of base clusterings, where the vertical error bar indicates the standard deviation over $20$ repetitions.  }
	\label{fig:M}
\end{figure}

\begin{figure*}[!t]
	\begin{minipage}[b]{0.16\linewidth}
		\centering
		\centerline{\epsfig{figure=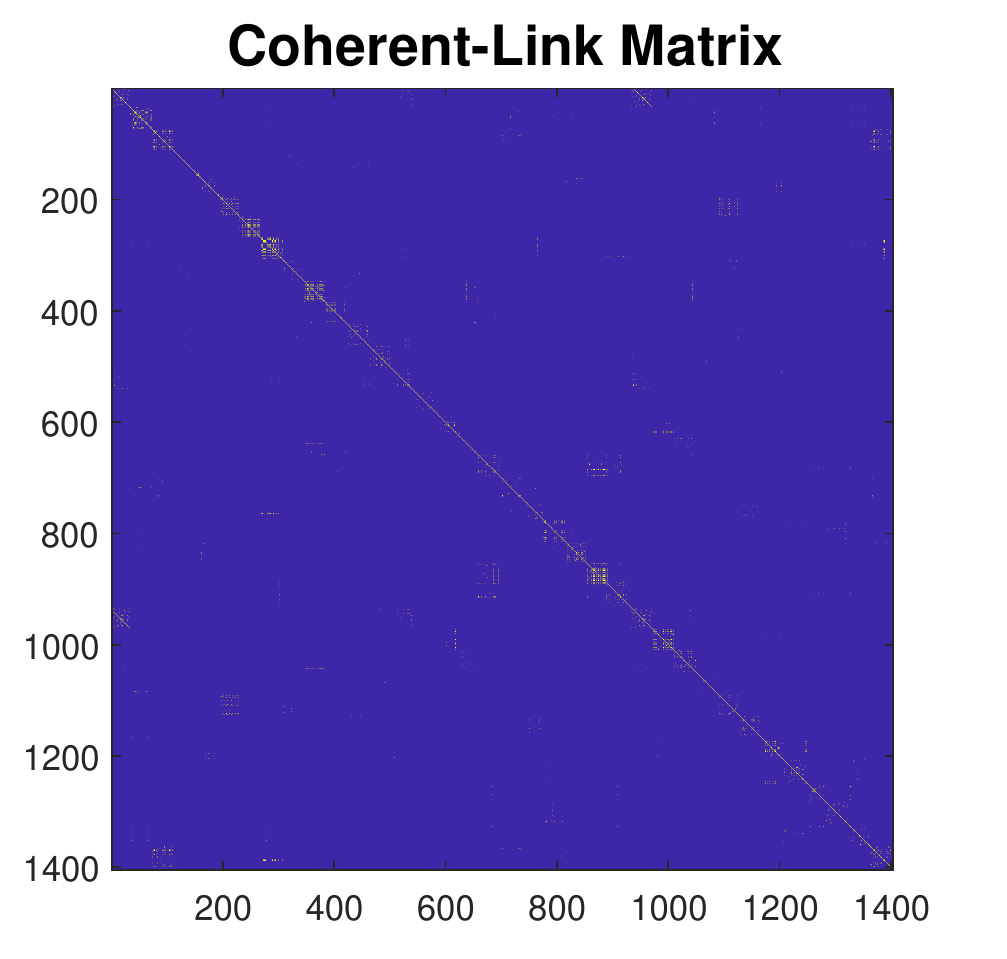,width=2.8cm}}
	\end{minipage}
	\begin{minipage}[b]{0.16\linewidth}
		\centering
		\centerline{\epsfig{figure=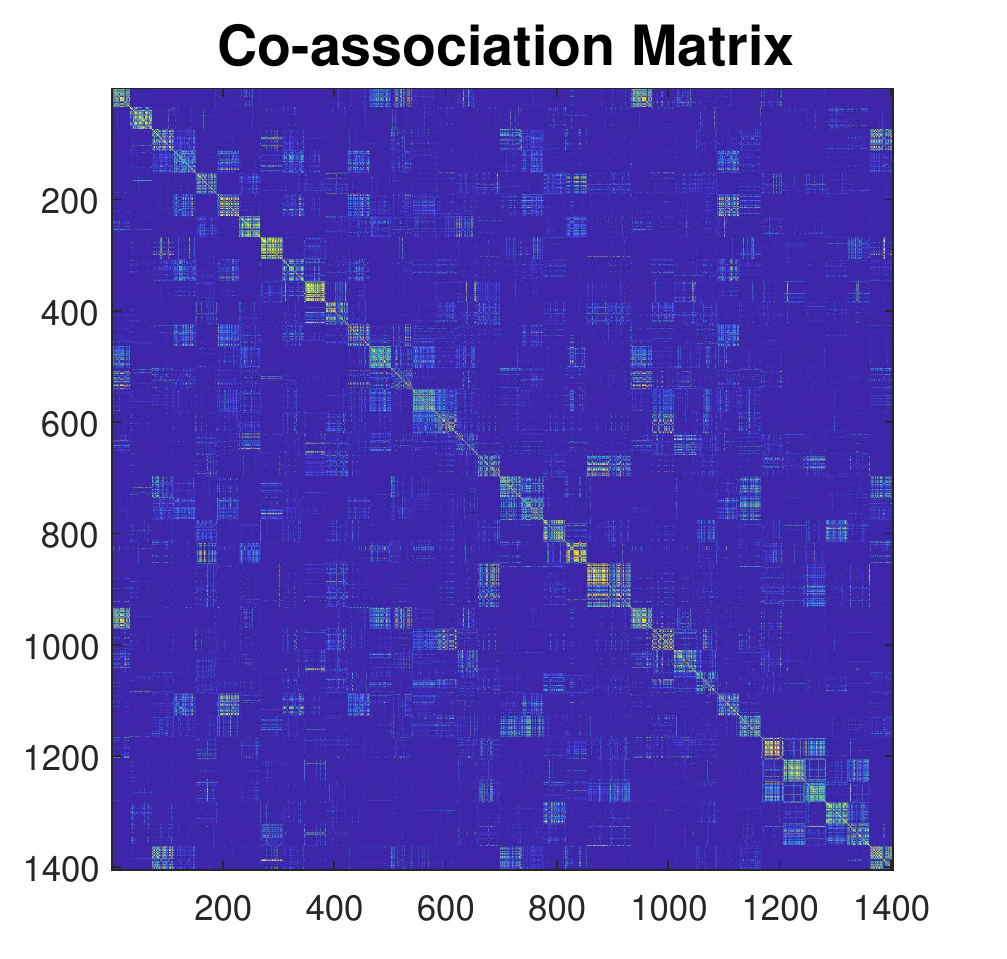,width=2.8cm}}
	\end{minipage}
	\begin{minipage}[b]{0.16\linewidth}
		\centering
		\centerline{\epsfig{figure=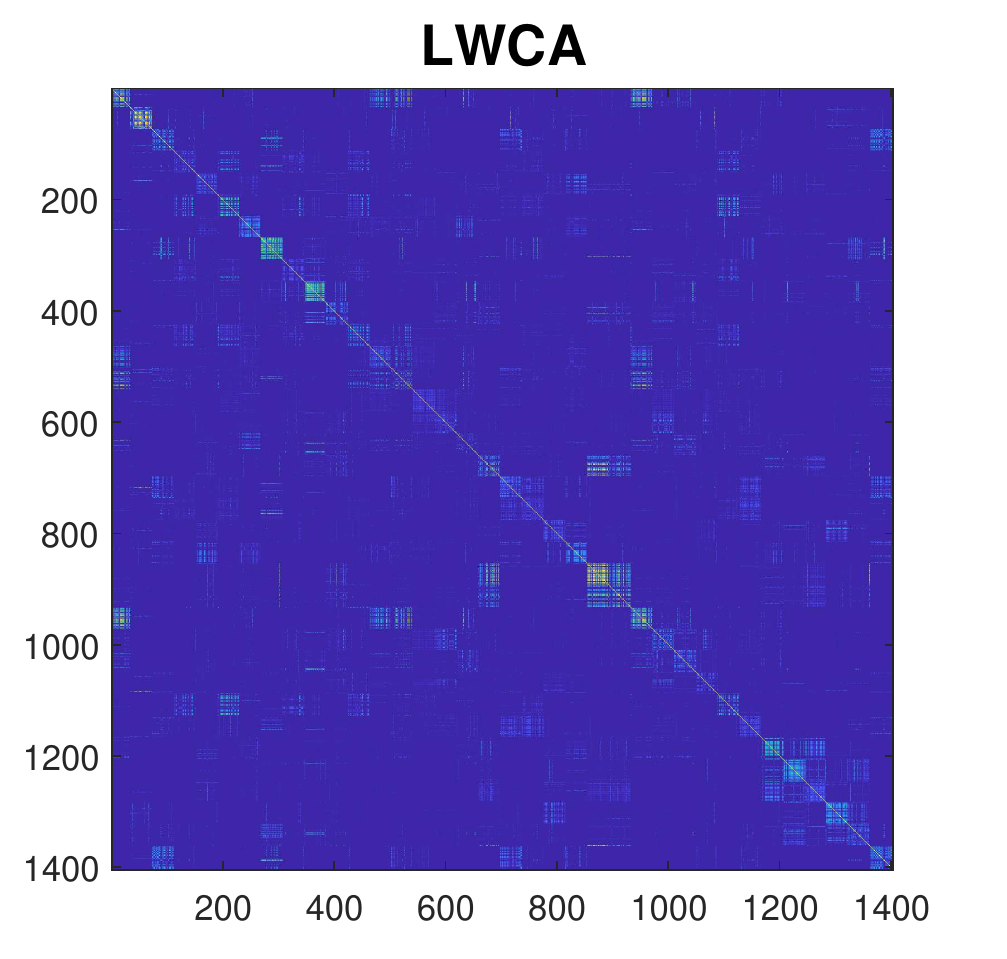,width=2.8cm}}
	\end{minipage}
	\begin{minipage}[b]{0.16\linewidth}
		\centering
		\centerline{\epsfig{figure=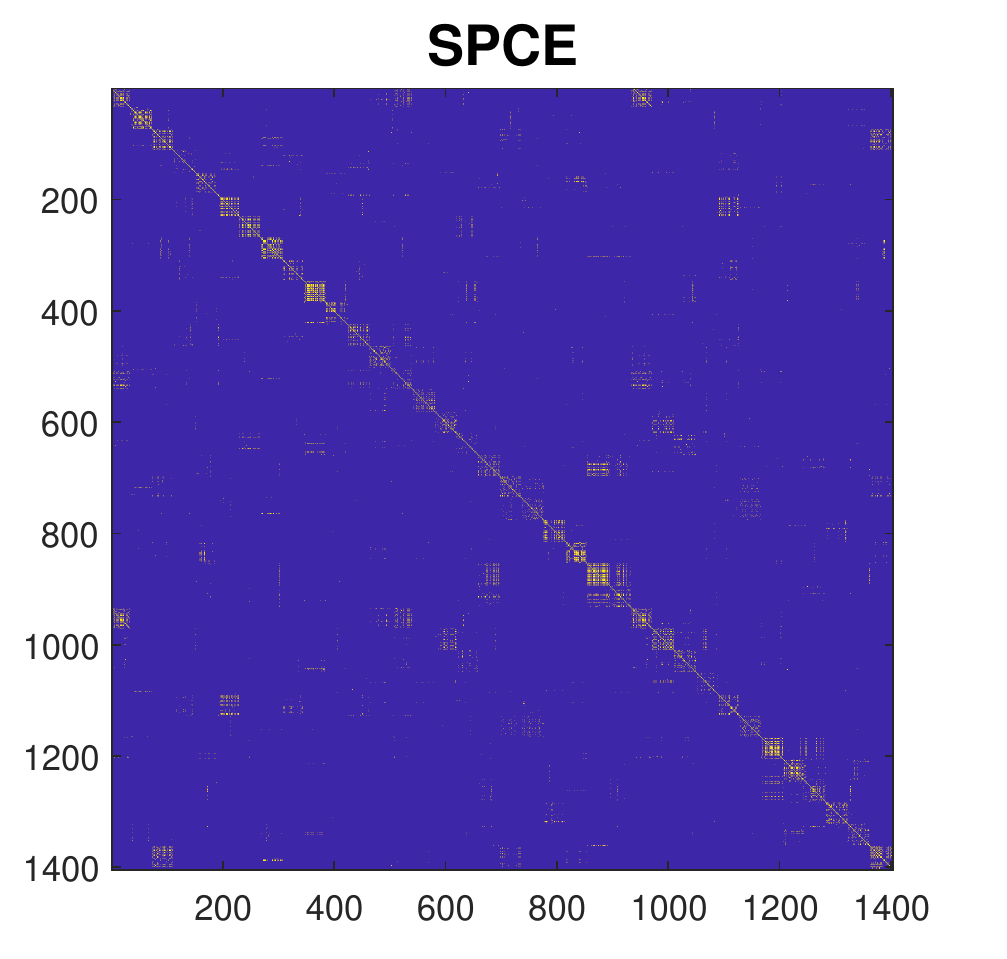,width=2.8cm}}
	\end{minipage}
	\begin{minipage}[b]{0.162\linewidth}
		\centering
		\centerline{\epsfig{figure=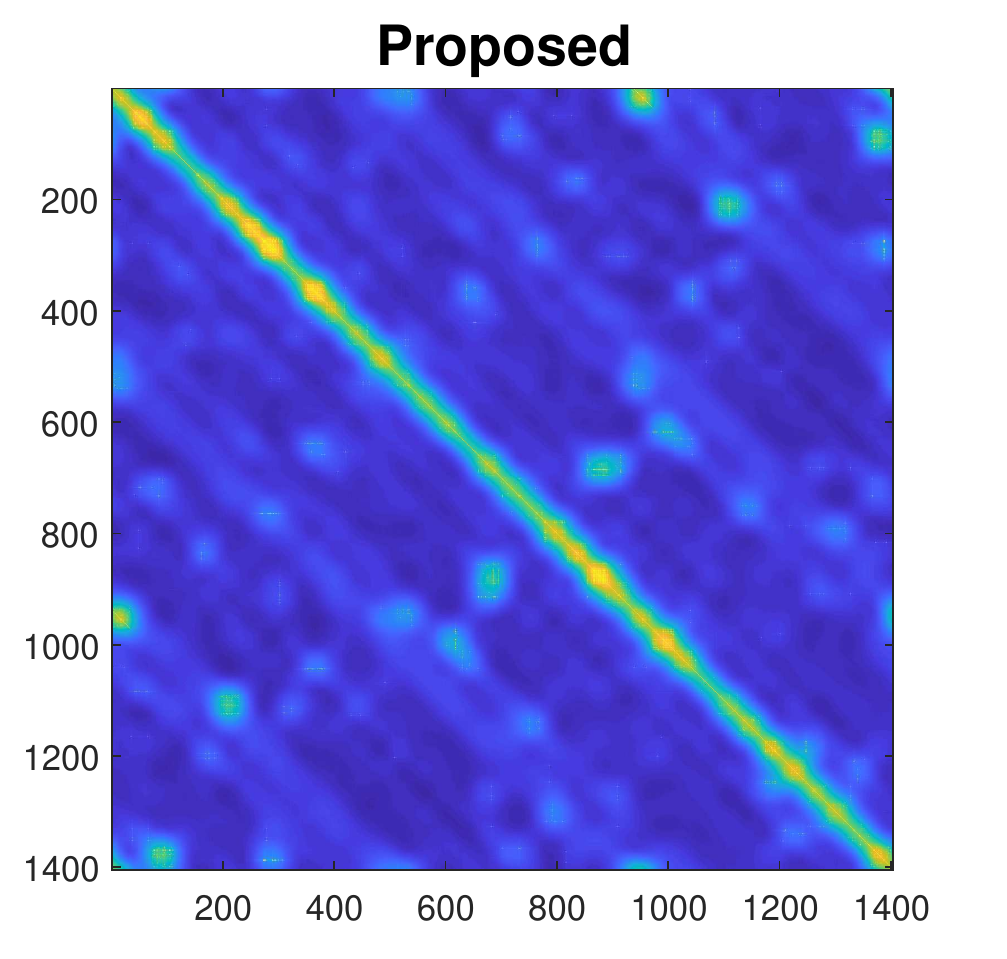,width=2.8cm}}
	\end{minipage}
\begin{minipage}[b]{0.162\linewidth}
	\centering
	\centerline{\epsfig{figure=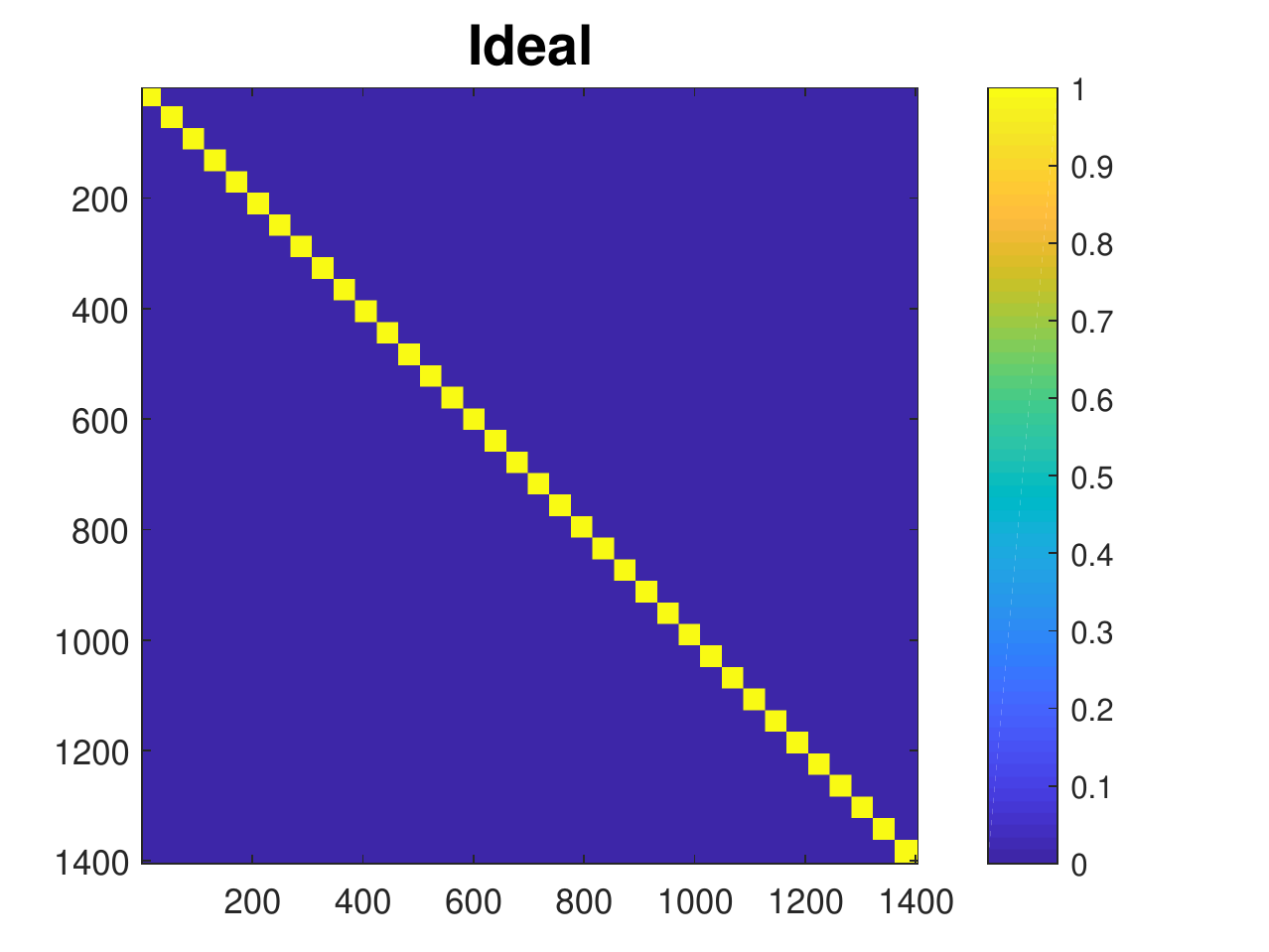,width=3.54cm}}
\end{minipage}
		\caption{Visual comparison of the learned pairwise similarity  matrices for different methods. All the matrices share the same color bar, and the brighter color indicates a larger value.} 
		\label{fig:latent-visual}
	\end{figure*}
	Tables \ref{table-BA}-\ref{ISOLET} show the clustering performance of all the methods over $7$ data sets, where we have the following observations. 
	First, the proposed methods including both Ours-EA and Ours-SC almost always outperform  all the compared methods under various metrics, which proves the universality of  the refined co-association matrix of the proposed model to different clustering methods. Moreover, Ours-SC usually performs better than Ours-EA, which means the refined co-association is more suitable for spectral clustering. Second, 
	the improvements of the proposed methods are significant. For example, on BinAlpha, compared with the best method under comparison,  Ours-SC increases the ACC from $0.454$ to $0.858$. On CalTech20, the highest ACC of the compared methods is $0.495$, while the ACC of Ours-EA is $0.726$. The improvements of  the proposed methods in terms of other metrics are also significant. Moreover, 
	the performance of Ours-SC on MF, MNIST, Semeion, Texture  are extremely good, i.e., all the metrics are quite close to $1$. 
	Those phenomena suggest that the proposed model brings a breakthrough in clustering ensemble. 
	Third, the highly competitive performance of the proposed model is achieved with a fixed hyper-parameter, proving the practicability of the proposed model. Besides,  the proposed model is also robust to different data sets, as both Ours-EA and Ours-SC consistently produce superior clustering performance on all the data sets. 
	\subsubsection{Comparison Against Base Clusterings.}
	We compared the average NMI of the our methods with that of all the base clusterings from the candidate clustering pool in Fig. \ref{fig:BPnmi}. It is clear that, on all the data sets, both Ours-SC and Ours-EA can significantly improve the NMI of the base clusterings, and Ours-SC outperforms Ours-EA in the majority cases.
	\subsubsection{Sensitivity to Hyper-parameter.}
	Fig. \ref{fig:lambda} shows the NMI of the proposed methods with different $\lambda$  on all the data sets, where we can conclude that: first,  a smaller $\lambda$ usually  leads to better clustering performance for both Ours-EA and Ours-SC, which demonstrates the importance of removing the incorrect connections   from the original  co-association matrix; and second,  for the majority data sets, the highest NMI occurs when $\lambda=0.002$ for both Ours-EA and Ours-SC, which proves the highly robustness of the proposed model to different data sets.
	\subsubsection{Performance with Different Number of Base Clusterings.} 
	Fig. \ref{fig:M} illustrates the influence of different numbers of the base clusterings to the proposed model, where we have the following observations. First, with the increase of the number of base clusterings, the NMIs of both  Ours-EA and Ours-SC generally increase, indicating that more base clustering are beneficial to the clustering performance. Second, with more base clusterings, the standard deviations generally become smaller for all the data sets, which suggests that more base clusterings can enhance the stability our methods. Third, for the majority data sets, $20$ base clusterings are sufficient for our methods to generate high value of NMI.
	\subsubsection{Comparison of the Learned Pairwise Similarity  Matrix.}
	Fig. \ref{fig:latent-visual} presents the cohere-link matrix, the traditional co-association matrix, the learned co-association matrices by LWCA \cite{huang2017locally}, SPCE \cite{zhou2020self} and the proposed model, and the ideal affinity matrix of BinAlpha, where all the matrices are normalized to $[0,1]$ and share the same color bar.
	Form \ref{fig:latent-visual}, we can observe that the coherent-link matrix is sparse, but its majority connections are correct, while on the contrary, the co-association matrix is dense, but with many incorrect connections in it. By exploiting the the low-rankness of the $3$-D tensor stacked by the coherent-link matrix and the association matrix, the refined  co-association matrix of the proposed model is quite close to the ideal one. Although there are some error corrections in it, almost all the relationships of two samples belonging to the same cluster have been correctly recovered, leading to high clustering performance. In contrast, there are many incorrect connections, but without enough correct connections in both the affinity matrices of LWCA and SPCE, which explains why they produced  inferior clustering performance than the proposed model. 
	\section{Conclusion}
	As the first work, we introduced low-rank tensor approximation to clustering ensemble.  Different from previous methods, the proposed model solves clustering ensemble from a global perspective, i.e., exploiting the low-rankness of a $3$-D tensor formed by the coherent-link matrix and the co-association matrix, such that the valuable information of the coherent-link matrix can be effectively propagated  to the co-association matrix. 
	Extensive experiments have shown that $i$), the proposed model improves current state-of-the-art  performance of clustering ensemble to a new level; $ii$), the recommended value for the hyper-parameter of the proposed model is robust to different data sets;  and $iii)$,  only a few base clusterings are required to generate high clustering performance.  
\bibliographystyle{IEEEtran}
\bibliography{aaai-2021-bib,bib}
\end{document}